# GIScience in the Era of Artificial Intelligence: A Research Agenda Towards Autonomous GIS

Zhenlong Li[1]*, Huan Ning[1]*, Song Gao[2], Krzysztof Janowicz[3], Wenwen Li[4], Samantha T. Arundel[5], Chaowei Yang[6], Budhendra Bhaduri[7], Shaowen Wang[8], A-Xing Zhu[2], Mark Gahegan[9], Shashi Shekhar[10], Xinyue Ye[11], Grant McKenzie[12], Guido Cervone[13], Michael E. Hodgson[14]

[1] Geoinformation and Big Data Research Lab, Department of Geography, The Pennsylvania State University, University Park, PA, USA

[2] Department of Geography, University of Wisconsin – Madison, WI, USA

[3] STKO Lab, Department of Geography and Regional Research, University of Vienna, Vienna, Austria

[4] Spatial Analysis Research Center, School of Geographical Sciences and Urban Planning, Arizona State University, AZ, USA

[5] Center of Excellence for Geospatial Information Science, U.S. Geological Survey, VA, USA

[6] NSF Spatiotemporal Innovation Center, Department of Geography & Geoinformation Science, George Mason University, VA, USA

[7] Science, Programs, and Partnerships, National Security Sciences, Oak Ridge National Laboratory (ORNL), TN, USA

[8] CyberGIS Center for Advanced Digital and Spatial Studies, Department of Geography and Geographic Information Science, University of Illinois Urbana-Champaign, IL, USA

[9] School of Computer Science, University of Auckland, New Zealand

[10] Department of Computer Science & Engineering, University of Minnesota, MN, USA

[11] Department of Landscape Architecture & Urban Planning and Center for Geospatial Sciences, Applications, & Technology, Texas A&M University, TX, USA

[12] Platial Analysis Lab, Department of Geography, McGill University, Quebec, Canada

[13] Institute for Computational and Data Sciences and Department of Geography, The Pennsylvania State University, University Park, PA, USA

[14] Department of Geography, University of South Carolina, SC, USA

*Leading first authors

Correspondence: zhenlong@psu.edu

## ABSTRACT

The advent of generative AI exemplified by large language models (LLMs) opens new ways to represent and compute geographic information and transcends the process of geographic knowledge production, driving geographic information systems (GIS) towards autonomous GIS. Leveraging LLMs as the decision core, autonomous GIS can independently generate and execute geoprocessing workflows to perform spatial analysis. In this vision paper, we further elaborate on the concept of autonomous GIS and present a conceptual framework that defines its five autonomous goals, five autonomous levels, five core functions, and three operational scales. We demonstrate how autonomous GIS could perform geospatial data retrieval, spatial analysis, and map making with four proof-of-concept GIS agents. We conclude by identifying critical challenges and future research directions, including fine-tuning and self-growing decision-cores, autonomous modeling, and examining the societal and practical implications of autonomous GIS. By establishing the groundwork for a paradigm shift in GIScience, this paper envisions a future where GIS moves beyond traditional workflows to autonomously reason, derive, innovate, and advance geospatial solutions to pressing global challenges. Meanwhile, as we design and deploy increasingly intelligent geospatial systems, we carry a responsibility to ensure they are developed in socially responsible ways, serve the public good, and support the continued value of human geographic insight in an AI-augmented future.

## 1 Introduction

Geographic Information Systems (GIS) have rapidly evolved over the past decades to enhance geospatial data collection, management, analysis, and visualization. GIS integrates data about locations (geographic data) with descriptive information (attribute data) to help people understand patterns, relationships, and trends in a spatial context (Goodchild, 1992). The evolution of GIS has been driven by successive waves of scientific advancement and technological innovation, including personal computers, the internet, high-performance and cloud computing (Yang et al., 2011), and cyberinfrastructure (Wang, 2010). At the same time, GIS and GIScience have always been influenced by its neighboring disciplines such as Cognitive Science and Computer Science. The emerging generative artificial intelligence is a type of artificial intelligence (AI) that can create new content such as text and images, by learning from large amounts of existing data (Feuerriegel et al., 2024). It has drawn tremendous attention throughout society (Chui et al., 2023) and the scientific community (Epstein et al., 2023), and has been studied in various fields, such as image creating, reasoning, writing, as well as programming (Sengar et al., 2024). Driven by generative AI, research has progressed toward the vision of Autonomous GIS as the next-generation GIS powered by AI and implemented as various GIS agents that can competently address geospatial problems with automatic spatial data collection, analysis, and visualization with minimal or no human intervention (Janowicz et al., 2020; Li & Ning, 2023). This section briefly introduces the history and development of GIS through its key stages driven by major disruptive technologies, as well as its anticipated evolution into autonomous GIS in the era of AI (Figure 1).

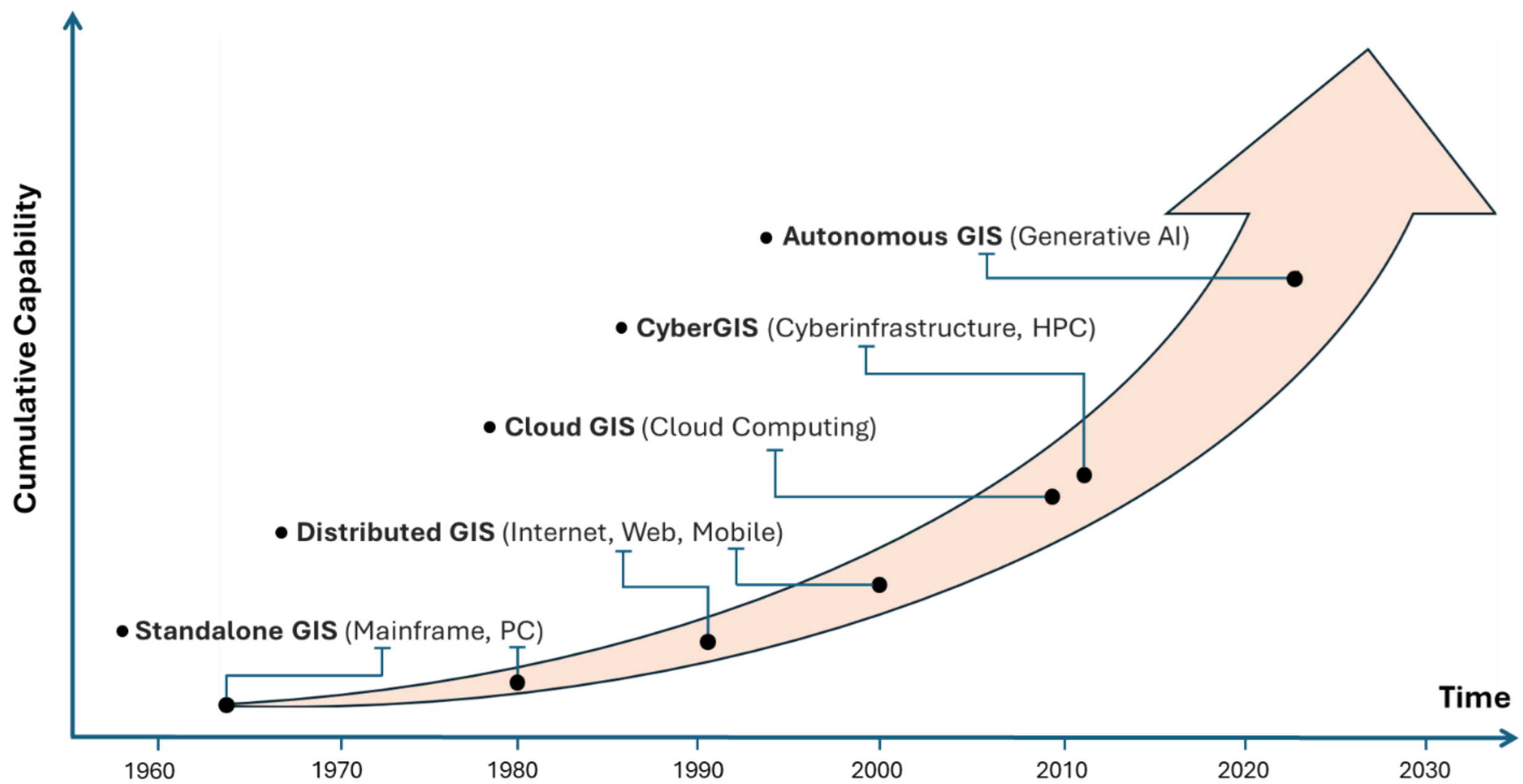

**Figure 1.** Evolution of GIS driven by major disruptive technologies.

### 1.1 Standalone GIS

The history of GIS dates back to the 1960s, with the development of the Canada GIS (CGIS) on a mainframe computer, which is often considered the first "true" GIS. CGIS, led by Tomlinson, was designed for land inventory and management. It introduced foundational concepts such as geospatial data layering and topology (Goodchild, 1992). In the 1970s, GIS software was limited to a few institutions, including CGIS, Harvard's Computer Graphics Lab, and the Oak Ridge National Laboratory. These systems ran mainly on mainframes, facing challenges related to computing power, geospatial data scarcity, and were only available to a very narrow range of users. The 1980s saw the emergence of commercial GIS solutions like ERDAS, Intergraph, and IDRISI, which expanded GIS accessibility to personal computers (PCs). The widespread adoption of PCs and the availability of (non-degraded) GPS in the late 20$^{th}$ century was a major turning point for GIS. The increasing computational power and accessibility of desktop computers popularized the development of desktop GIS, shifting spatial analysis from mainframe systems to personal workstations. Early systems such as ArcInfo, released in 1982 by Esri, used command-line tools for data management, analysis, and visualization to transform manual cartography into a digital process. Later software tools such as ArcGIS, QGIS, and GRASS GIS further reinforced the role of desktop GIS in applications ranging from cartographic design and geospatial modeling to geospatial data integration and analysis (Österman, 2014).

### 1.2 Distributed GIS

Distributed GIS emerged in the late 20thcentury as organizations sought to overcome the limitations of standalone GIS by leveraging networked environments. The term typically refers to a framework where spatial data, processing, and services are distributed across multiple interconnected systems rather than being confined to a single computer. Early forms of distributed GIS relied on local area networks (LANs) and client-server architectures, enabling users to access shared geospatial databases and perform remote spatial analysis. The rise of the Internet and World Wide Web enabled the development of distributed GIS into Spatial Data Infrastructures and WebGIS, which provides GIS functionalities through web browsers (Kearns et al., 2003). The first WebGIS application "Xerox PARC Map Viewer" was developed in the early 1990s (Putz, 1994). The open source MapServer (Lime, 2008) and Esri's commercial Arc Internet Map Server (ArcIMS), launched in the 1990s, are examples of early WebGIS platforms. With the prevalence of Web 2.0, characterized by increased user-web interaction and user-generated content (Wilson et al., 2011), WebGIS entered another stage by including interactive geo-visualization, (distributed) spatial analysis, and community mapping capabilities – often in the form of so-called Web Servies as defined by the Open Geospatial Consortium (OGC) (Chen et al., 2025). OpenStreetMap was founded in 2004 (Mooney & Minghini, 2017), and Google Maps was launched in 2005. The rapid growth of mobile devices and wireless communication technologies in the 21st century further drove the evolution of distributed GIS into Mobile GIS where GIS capabilities are integrated into smartphones, tablets, and other portable devices, enabling a wide range of location-based services (LBS) such as navigation applications (Huang, 2022). Distributed GIS extended the reach of GIS to global audiences and non-specialists, fundamentally changing how spatial data are shared and consumed.

### 1.3 Cloud GIS

Cloud GIS builds upon distributed GIS and cloud computing technologies by offering scalable and flexible GIS solutions hosted on cloud platforms (Yang et al., 2011). These systems eliminate the need for local infrastructure, reducing costs and maintenance efforts while providing global accessibility. Essentially, Cloud GIS provides GIS software as a service (SaaS) or infrastructure as a service (IaaS) (Xia, 2022). ArcGIS Online, launched in 2012, is a good example of Cloud GIS that offers cloud-based mapping and analysis capabilities (Esri, 2025). Another powerful Cloud GIS platform is Google Earth Engine, which runs on Google Cloud and supports large-scale geospatial applications such as vegetation monitoring, deforestation tracking, and wildfire mapping (Zhao et al., 2021). Governments can use Cloud GIS for disaster risk assessment, leveraging real-time data feeds to predict flood zones and monitor storm impacts (Albrecht et al., 2020). In the private sector, industries such as logistics use cloud GIS to optimize network planning and route efficiency (Khune, 2024). Cloud GIS also empowers education and outreach by providing accessible platforms for teaching spatial literacy and engaging communities in participatory mapping projects (Kholoshyn et al., 2019).

### 1.4 CyberGIS

CyberGIS emerged in response to the need for computationally intensive spatial modeling, analysis, and geovisualization. This paradigm integrates advanced cyberinfrastructure—high-performance, data-intensive computing, and networked systems—to tackle complex, large-scale geospatial problems (Wang, 2010). By leveraging scalable computing capabilities, CyberGIS allows researchers and users to analyze complex and massive geospatial data efficiently, enhancing applications such as flood modeling, agricultural monitoring, and urban growth across many spatial and temporal scales. Moreover, CyberGIS facilitates multi-disciplinary collaborations by integrating spatial analysis with data from disciplines such as climatology, economics, and epidemiology, and enhances decision-making processes for various stakeholders (Li et al., 2013). For example, disaster response agencies have used CyberGIS to simulate evacuation scenarios for policy and resource planning (Vandewalle et al., 2019), while operational platforms like Forwarn, developed by the USDA, enable near-real-time vegetation monitoring across the U.S. (Norman et al., 2013). The integration of machine learning within CyberGIS environments further enhances predictive capabilities, such as forecasting deforestation and urban heat island effects (Lyu et al., 2022).

### 1.5 Autonomous GIS

Since the first release of ChatGPT by OpenAI in November 2022, generative AI, particularly transformer-based deep neural networks exemplified by Large Language Models (LLMs), has drawn tremendous attention from the public, industry, and academia (Kaddour et al., 2023). These models, trained on massive corpora, exhibit remarkable abilities to understand input text and generate associated human-like text. More importantly, by providing context information, LLMs emulate reasoning processes, plan, and perform tasks across domains. Thus, an LLM can be used as a decision core to power applications to react and adapt to changing environments. These applications pave the way for autonomous agents (Li & Ning, 2023; Wang et al., 2024; Yang & Velazquez-Villarreal, 2024). Along with other generative AI techniques (Sengar et al., 2024), the applications of LLMs and multimodal LLMs have been broadly explored. The GIScience community has noticed the potential of generative AI and inquired into various applications (Fanfarillo et al., 2021; Wang et al., 2024).

We emphasize that the generative AI's abilities in reasoning, text and code generating, vision, and general knowledge replication are "general" for a GIS analyst. Generative AI powers autonomous agents that can make decisions based on environmental changes rather than relying on pre-defined rules or strategies. This decision-making ability brings enormous benefits for spatial analysis, which can be presented as a geoprocessing workflow. We believe that generative AI as well as its potential progression towards future artificial general intelligence (AGI) is a disruptive technology that paves the way for autonomous GIS.

We define autonomous GIS as AI-powered next-generation geographic information systems that leverage the generative AI's general abilities in natural language understanding, reasoning, and coding for addressing geospatial problems with automatic spatial data collection, analysis, and visualization with minimal or no human's intervention (Janowicz et al., 2020; Li & Ning, 2023; Zhu et al., 2021). Geospatial problem-solving can be framed as geoprocessing workflows, which consist of a series of spatial data operations such as data retrieval, reprojection, spatial analysis, modeling, geo-visualization, and report generation. Autonomous GIS leverages generative AI such as one or multiple LLMs as a decision core to plan, implement, and execute these workflows. This involves retrieving or collecting spatial data from existing repositories or sensors, utilizing existing GIS tools or developing new ones as needed, and processing the data to produce final outputs such as maps, charts, and reports. This capability is based on yet another step in the history of GIS, namely the push to open-source libraries and application programming interfaces (APIs) instead of all-in-one systems.

It is important to recognize that the evolution of GIS, from desktop systems to CyberGIS, does not follow a linear path where newer technologies replace older ones. Instead, each new form of GIS emerged alongside technological advancements, addressing specific needs and expanding the overall ecosystem of geospatial technologies (Figure 1). Desktop GIS remains essential for many individual and specialized tasks. At the same time, WebGIS brings geospatial capabilities to a global audience via browsers, Cloud GIS integrates scalability and flexibility into service-oriented geoprocessing workflows, and CyberGIS tackles computationally intensive analyses and enables collaborative problem solving. These modalities coexist and intersect in many ways, complementing each other in addressing numerous applications' various and growing demands.

In this vision paper, we aim to use clear and accessible language rather than technical jargons to ensure that a broad audience can engage in the ideas presented. Our goal is to inspire more researchers and practitioners to participate in and shape this emerging paradigm of autonomous GIS. Additionally, we deliberately refrain from detailing specific technologies or developments in generative AI, recognizing that this field is rapidly evolving, with breakthroughs continuously reshaping the research landscape. Instead, our vision focuses on fundamental goals, core concepts, and critical challenges that may define the trajectory of autonomous GIS.

## 2 Autonomous GIS: the next-generation AI-powered GIS

One of the key impacts of generative AI and its potential progression towards artificial general intelligence is the automation of digital operations from both the perspectives of accessibility and intelligence, evolving GIS into autonomous GIS. In this section, we present our vision for autonomous GIS by expanding and deepening the previous discussions by Li and Ning (2023), focusing on automation goals, levels of autonomy, and application scales. This section sets the stage for discussing the current research landscape, challenges, and research agenda of autonomous GIS in the following sections.

### 2.1 Conceptual framework of autonomous GIS

We believe that autonomous GIS represents an emerging paradigm of integrating AI with GIS, where it is not merely another tool but becomes an "artificial geospatial analyst" or "digital agent" who knows how to use GIS tools and geographical analysis and with what data to solve geospatial problems. In this context, "agent" refers to an autonomous software entity that can take requests and perceive its environment, make decisions, formulate plans, and take actions. Instead of being a specific system, we view autonomous GIS as a broad concept and a distinct paradigm similar to WebGIS and Cloud GIS. Compared to traditional GIS systems, autonomous GIS is intended to improve accessibility, aiming to democratize spatial analysis and geospatial technologies for a wider audience.

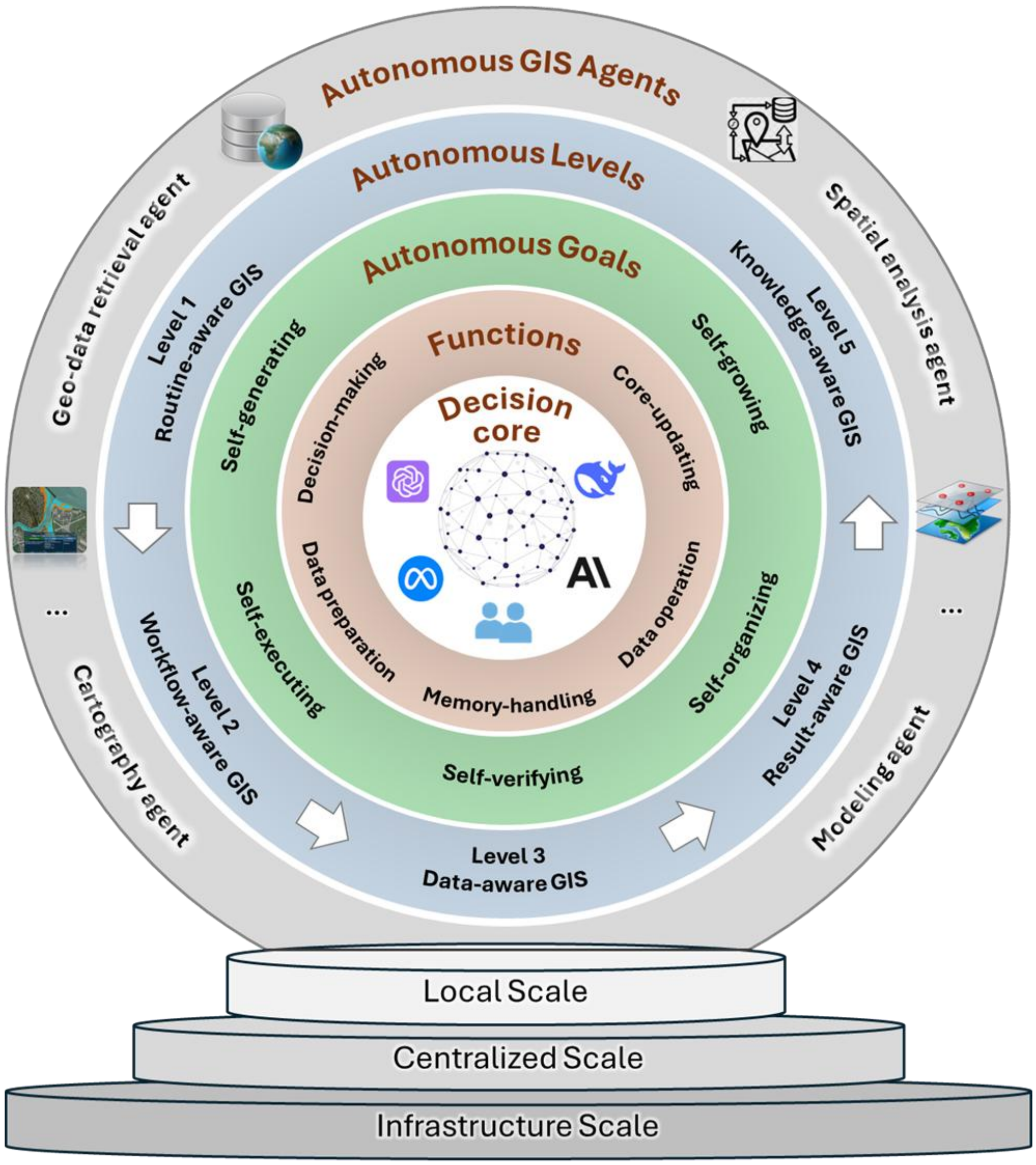

**Figure 2.** Conceptual framework of autonomous GIS

We envision that autonomous GIS should be designed as a programming- and data-centric framework that automates coding to address geospatial problems. This framework can be implemented into various "autonomous GIS agents" that work individually or collaboratively. These agents assist in and automate various spatial tasks, such as preparing data, designing geoprocessing workflows, performing spatial analyses, conducting predictive modeling, extracting insights, creating maps, evaluating the results, adjusting workflows, generating reports, and making recommendations. An autonomous GIS agent does not need to be all-round and versatile; it can be specialized to focus on a specific task, such as data collection or cartography, being built upon exiting GIS components. Specialized agents can collaborate to accomplish complex tasks, much like human teams. We consider these GIS agents as the building blocks of autonomous GIS. In this sense, autonomous GIS can also be referred to as "agent-based GIS" or "agentic GIS".

From a user's perspective, autonomous GIS functions as a digital agent or a set of agents that receive geospatial problem queries via natural language rather than specific commands. These agents serve as the primary user interface for interacting with the system. For example, a user might ask, "How do severe weather events impact the frequency and distribution of traffic accidents in the U.S.?" The GIS agent(s) will collect publicly available weather and traffic accident data, test multiple models to explore the relationship and adjust the model parameters through an iterative process based on its review of modeling results. By comparing and synthesizing the results from multiple models, the GIS agent (s) generates a reliable and reproducible conclusion and presents the user a report that includes maps, charts, result descriptions, interpretations, and recommendations. In addition, autonomous GIS could also power robots in the physical world that are designed to map and navigate through various changing physical environments, such as autonomous vehicles (Garikapati & Shetiya, 2024), vacuum robots (Kim et al., 2019), and drones (Hanover et al., 2024).

From an implementation perspective, autonomous GIS comprises five key functions, including *decision-making*, *data preparation*, *data operation*, *memory-handling*, and *core-updating* (Figure 2). *Decision-making* function generates geoprocessing workflows and manages tasks. *Data preparation* ensures that the system autonomously gathers and selects geospatial data, addressing uncertainties in data coverage, quality, and compliance with regulations. *Data operation* handles spatial data processing tasks, such as transformation and visualization, with built-in error management. The *memory-handling function* logs workflows, results, and new knowledge to support future use and improvements. *Core-updating* function enables self-learning by fine-tuning the AI core with past successes and failures, ensuring continuous enhancement. These functions, powered by AI and pre-defined strategies, allow autonomous GIS to execute tasks with minimal human intervention, adapt to new challenges, and refine its capabilities over time. Not every autonomous GIS implementation or agent must realize all these functions, as the user's demands vary. For example, the agent with *decision-making* and *data operation* functions can serve most simple spatial analyses of multiple steps, such as watershed generation based on provided digital elevation models (DEM) and study area boundaries.

Li and Ning (2023) identified five core goals for autonomous GIS to operate with minimal reliance on human analysts: self-generating, self-organizing, self-verifying, self-executing, and self-growing. (Figure 2). This paper further elaborates on the rationale behind these goals (Section 2.2). Whereas autonomous GIS aims to minimize human intervention, we anticipate its development will follow an incremental approach, with varying levels of autonomy, similar to the staged advancements seen in self-driving vehicles. To this end, we categorize autonomous GIS into five autonomous levels, defined by the system's ability to manage uncertainty, providing a structured pathway toward full autonomy (Section 2.3). Based on the computing resources available to the agents, we further categorize autonomous GIS into three scales: local, centralized, and infrastructure (Section 2.4).

As we conceptualize autonomous GIS as an artificial geospatial analyst based on its technological capabilities, its level of autonomy reflects the degree to which it can independently conduct spatial analysis without human technical intervention. A highly autonomous GIS requires minimal technical support but does not preclude frequent interactions with humans, as it must gather sufficient context and task requirements to operate correctly and effectively. When humans intervene because the system is heading in the wrong direction, we classify this as interaction if the issue stems from incomplete or inaccurate contextual information. However, if the disruption occurs despite having all the necessary context and is due to the system's technological limitations, we consider it an intervention.

## 2.2 Goals of autonomous GIS

We envision that autonomous GIS should achieve five autonomous goals from the perspective of system behavior: self-generating, self-executing, self-verifying, self-organizing, and self-growing. Note that these goals are qualitative and descriptive, emphasizing the functional characteristics of an autonomous GIS rather than rigid technical specifications.

### 2.2.1 Self-generating

Self-generating involves creating hypotheses, research ideas, geoprocessing workflows, code, and insights. For data operations tasks, autonomous GIS should be capable of generating data processing workflows and corresponding programs to execute them. Regarding research or modeling tasks, we hope autonomous GIS will be able to generate innovative ideas (Hu et al., 2024; Meincke et al., 2024), providing feasible, novel, or cost-effective solutions for geospatial problems. Once research hypotheses and ideas are established, autonomous GIS should generate geoprocessing workflows and code to test or validate them. These workflows form the foundation of spatial analysis, with code-based scripts or programs serving as their practical implementations. Previous studies have demonstrated that current state-of-the-art LLMs can interpret both data processing and research tasks expressed in natural language and translate them into workflows and executable code. Beyond generating workflows and code, a crucial next step is deriving meaningful insights from the results. Although there are some early attempts in GIS report writing with LLMs (Starace & Di Martino, 2024), we believe that effective reporting relies on accurate and reasonable interpretation of results, an area that remains largely unexplored. Many multimodal LLMs can accurately process visual information, benefiting the insight generation as geovisulization is critical for humans to obtain insights. We believe the self-generating goal is both practical and achievable, and it should be prioritized as a foundational step in developing autonomous GIS.

#### 2.2.2 Self-executing

Autonomous GIS should be able to act, achieving the goal of self-executing. This involves preparing data, executing geoprocessing workflows, and ultimately converting geographic data into results containing numbers, tables, or maps using given tools and resources. To achieve this, an appropriate hardware and software environment with sufficient documentation is needed to execute solutions for the given tasks. Python is a commonly used environment where the system executes the generated code in the Python runtime to access data from storage, process it, and produce results. Another approach to self-execution involves simulating computer use, where the agent operates the GIS software such as ArcGIS (a commercial GIS) or QGIS (an open-source GIS) through screen-based interactions (OpenAI, 2025). Although still in its early stages, this approach is similar to the behavior of a geospatial analyst, allowing the agent to directly use extensive existing geospatial tools, such as shortest path search or database management.

#### 2.2.3 Self-verifying

The goal of self-verifying requires the system to validate the results of its actions, including data operations at each step and the final outputs, ensuring that each operation performs as expected. For example, the system should confirm whether the loaded data are accurate, the spatial join is successful, and the aggregation results are reasonable. Achieving this goal involves several key components, including 1) geoprocessing workflow review: check if the generated workflow is reasonable regarding the task context and the data condition; and 2) step verification: inspect the results of each step to ensure their validity and mitigate the uncertainty propagation in the geoprocessing workflows. Stepwise verification is crucial, as the accuracy of the outcome relies on the correctness of each intermediate step. This raises critical questions: What constitutes "correctness"? What criteria should be applied to attributes, tables, images, and visualizations? At what level of accuracy should results be considered acceptable? What are the potential consequences if an erroneous result is passed to the next step? Considering the varied data modalities, sources, qualities, and task contexts, we believe one critical task to achieve the self-verifying goal is to define clear and practical criteria to address uncertainty. It is important to note that the verification discussed here refers to the correctness and reasonableness of data operations rather than their alignment with ground truths.

#### 2.2.4 Self-organizing

We also expect autonomous GIS to self-organize. Like humans and other systems, agents may have limited resources for problem-solving and may benefit from carefully managing and organizing these resources. The system should know how to allocate resources and actions to address tasks effectively, including time, computing power, memory, and communication with users. For example, it will notify users when there are insufficient resources to accomplish the assigned tasks. It should also label successful and failed attempts to build experience and avoid repeating similar mistakes in the future. These practical constraints require strong project management skills. Hence, one type of autonomous GIS agent could act like a qualified project manager, capable of organizing resources, coordinating actions, and collaborating with other specialized GIS agents to organize a “team” to handle complex tasks. Such an agent also needs to recognize when to request additional context from human users or collaborate to clarify goals, resolve issues, or generate new ideas. Overall, the goal of self-organization reflects the system’s ability to manage external resources and internally generated outputs, ensuring that tasks are completed efficiently and effectively. Given the potentially very high resource requirements, this also means that GIS agents should be able to anticipate their footprint, e.g., in terms of utilized energy, carbon emissions, and other aspects of AI sustainability (Shi et al., 2025) .

#### 2.2.5 Self-growing

Autonomous GIS is expected to self-grow by continuously learning from past tasks, including successful and failed attempts and external documents. This means it should be able to improve future spatial analysis and modeling by summarizing general rules and the reasoning behind its actions, identifying the needed knowledge or skills, and updating its strategies. Among the five goals, self-growing is the most challenging to achieve. It requires the system to perform tasks more efficiently over time—faster, with greater accuracy, or by consuming fewer resources. It also needs to adjust its capability to meet the changing mission of the user and organization. Recent research (Li & Ning, 2023; Ning et al., 2025) highlights that even the most advanced generative AI models, such as GPT-4o, still struggle with spatial programming tasks due to a lack of background knowledge or domain-specific skills, such as handling map projections or performing spatial join operations.

Imagine that a generative AI model is like a newly graduated college student who needs to learn necessary knowledge and skills through continuous application. After several years, this person becomes an experienced contributor to their field of application. Currently, most AI models cannot accumulate experience; they come to the workplace like fresh employees every day, doing some tasks without any improvements. The capabilities of generative AI models stagnated after their creation, and addressing these weaknesses remains a significant challenge that will require technologies such as retrieval-augmented generation (Lewis et al., 2020) and knowledge graphs (Kommineni et al., 2024). We encourage the geospatial community to explore ways to enable autonomous GIS agents to self-grow and incorporate external knowledge bases into their processes. Potential approaches include fine-tuning models using techniques such as Low-Rank Adaptation or LoRA (Hu et al., 2021), which helps the system adapt and improve its capabilities over time; it adds small, trainable layers to the model, allowing efficient adaptation without changing the original model's core weights.

### 2.3 Levels of autonomous GIS

It is natural to define incremental levels for autonomous GIS, drawing inspiration from the levels of autonomous vehicles and the autonomation community (Antti Karjalainen, 2024; Cook, 2024; Greyling, 2024; Y. Huang, 2024; Morris et al., 2024; Synopsys, 2019). We categorize autonomous GIS into five levels, ranging from Level 1 to Level 5 (Figure 3), with Level 0 referring to no automation, where humans need to manually conduct all data processing and analyses in a GIS. Note that this paper focuses exclusively on technological autonomy, where human intervention refers to technological support or correction rather than supplementing or clarifying context or task requirements.

Level 1 is *routine-aware* GIS, which can conduct the manually defined geoprocessing workflows, such as ArcGIS Model Builder and QGIS Model Designer. Level 2 is *workflow-aware* GIS, which can generate workflows autonomously to accomplish spatial tasks according to the given data and expected results, including implementing selected models for specific analyses. Level 3 is *data-aware* GIS, which emphasizes the ability to select existing datasets for the tasks or collect new ones when needed. *Result-aware* GIS, Level 4, underlines the ability to understand the results and to improve workflow and data usage accordingly with limited human intervention. Finally, Level 5 represents *knowledge-aware* GIS, a fully autonomous system capable of performing spatial analyses without human supervision. It develops general rules, refines strategies, and improves geoprocessing workflows to achieve the self-growing goal by learning from its successful and failed attempts.

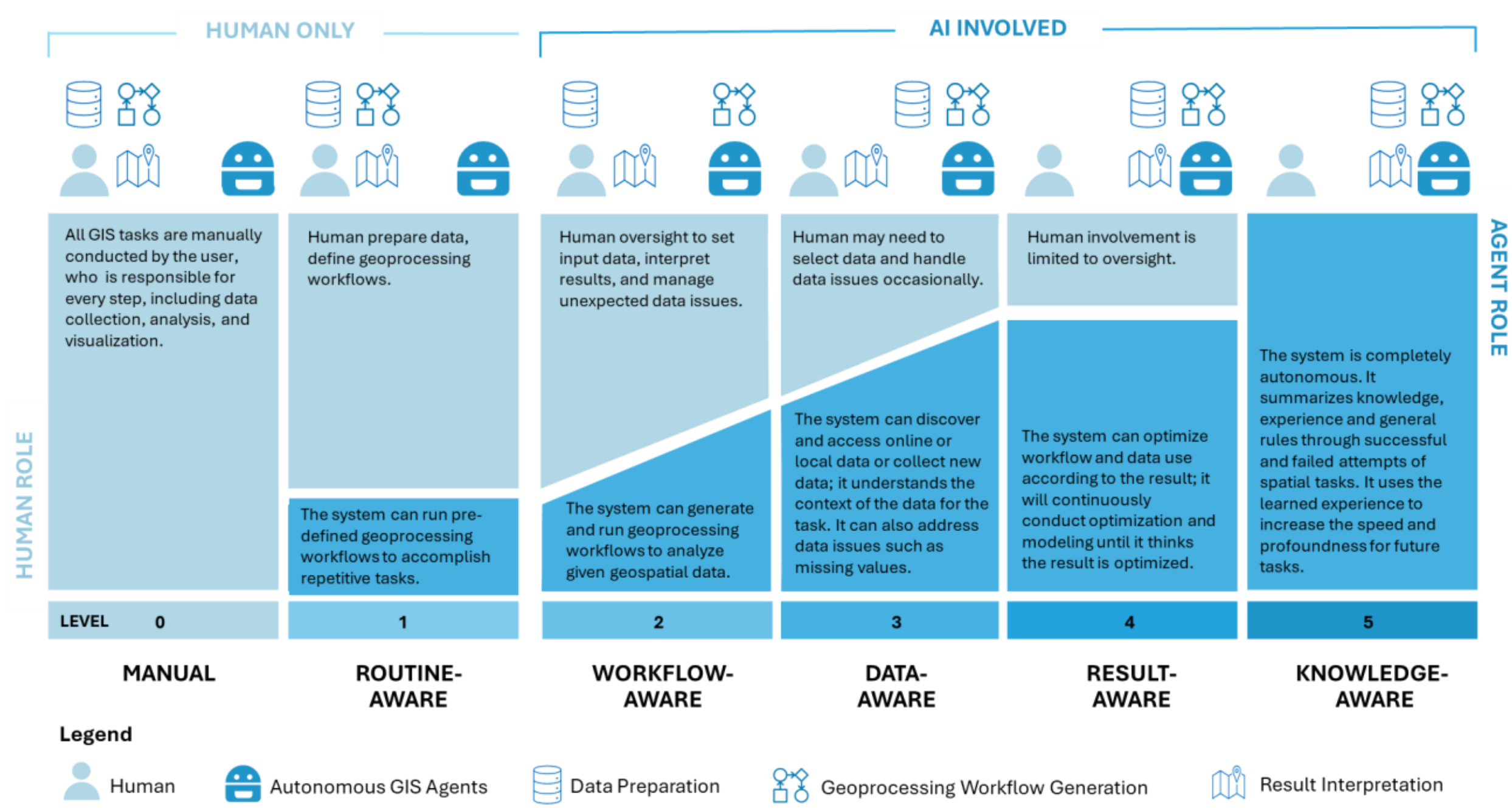

**Figure 3**. Levels of autonomous GIS, inspired by Mike Lemanski (Smith, 2016)

### 2.3.1 Level 0: Manual GIS (No Automation)

Manual GIS represents a fully manual system where all GIS tasks, including data collection, analysis, and visualization, are manually determined and executed by the user using a GIS without any automation (aside of cases such as model builders, scripting, and batch operations). The system only provides data manipulation and visualization tools, requiring the user to handle every aspect, including identifying and correcting errors. For instance, an analyst uses GIS to manually import census data and boundary files, calculate population density, and overlay them on a map to create a heatmap.

### 2.3.2 Level 1: Routine-aware GIS (Process Automation, human driven)

Routine-aware GIS introduces basic process automation for pre-defined repetitive tasks, such as data import, transformation, and spatial functions such as buffering or clipping. After the user defines workflows and parameters, the system automates their execution, such as data batch processing. It also handles basic errors, although complex corrections still require human intervention. For example, the system could automatically load and reproject shapefiles of a city's road network and clip them to specific boundaries, with the user setting up the geoprocessing workflow using tools such as ArcGIS Model Builder or programming language such as Python. Humans drive analyses using routine-aware GIS.

### 2.3.3 Level 2: Workflow-aware GIS (Workflow Generation, human assisted)

At this level, AI may begin to play an important role by expanding automation to include the generation and execution of geoprocessing workflows for given tasks, making traditional GIS systems into workflow-aware GIS. It can select and apply appropriate GIS tools and models to address geographic phenomena based on the data prepared by users (humans or agents). However, human involvement remains essential for providing input, interpreting results, and managing data-related issues such as missing data and data inconsistencies as the level's adaptability to unexpected execution challenges is limited. For example, a workflow-aware GIS might autonomously create a geoprocessing workflow for mapping flood-prone areas using terrain and satellite data, execute hydrological models, and produce a flood risk map, with humans overseeing or refining the process as required.

### 2.3.4 Level 3: Data-aware GIS (Data Preparation Automation, human supervised)

Data-aware GIS integrates data selection and contextual understanding, automating data collection and preparation based on tasks provided by humans and agents. It represents the autonomous capability to handle uncertainty in data preparation (section 2.3.7) rather than serving as a prerequisite for geoprocessing workflows or the *workflow-aware* GIS. The system can recognize the applicability and drawbacks of spatial data types, handle inconsistencies such as missing values and incomplete coverages, and adapt classification methods to accommodate varying spatiotemporal resolutions. Humans will still need to supervise the system, guiding data selection and use. For example, when mapping green spaces, the system can identify variations in satellite imagery resolution, apply suitable classification techniques, and flag low-quality data for review. Human analysts should ensure that the data, modeling, results, and result interpretations are reasonable.

### 2.3.5 Level 4: Result-aware GIS (Result Interpretation, minimal supervision)

Result-aware GIS introduces optimization based on result assessments, allowing the system to adjust workflows, models, and data use to improve outcomes. The decision-core understands potential causal relationships between the input, model, and output, so it can alter the input and modeling to achieve the desired output. Result-aware GIS can conduct iterative modeling, as most modeling processes are iterative. The system can adjust the data, model, model parameters, and workflows to produce optimized results. Human involvement may become minimal, primarily limited to oversight. For example, when predicting traffic congestion, the system uses real-time sensor data and autonomously optimizes the workflow; when realizing the prediction modeling takes too much time, it drops less critical sensors to enhance processing speed.

### 2.3.6 Level 5: Knowledge-aware GIS (Self-growing, no supervision)

Knowledge-aware GIS represents a fully autonomous system capable of learning from previous tasks and external resources to improve its future performance. Like humans, it can accumulate experience from practice. The trial-and-error process is the key; it can identify and summarize the general rules and knowledge behind the rationale of successful and failed tasks. It can also update its decision core not only based on learned rules, knowledge, but also external documents. It independently conducts predictive or explanatory analyses, continuously improving its speed and accuracy. For example, the system autonomously analyzes historical climate data and real-time weather patterns to predict future flood risks. It tests multiple models and various data inputs, evaluates the outcomes, and produces reasonable results. Through this process, the system learns the applicability of various models and datasets, enabling it to apply the most effective strategies for similar flood predictions in the future.

### 2.3.7 Uncertainty and autonomous levels

We define the autonomous levels by considering the agent’s ability to address uncertainty. We expect autonomous GIS agents to make reasonable decisions and handle unseen situations — a core aspect of handling uncertainty. Spatial analysis involves various levels of uncertainty across the stages of data preparation, geoprocessing workflow construction, and result evaluation. Workflow (data preparation excluded) uncertainty is the least among the three stages. It is like a “blueprint,” and its patterns are highly structured, often represented as a directed acyclic graph, with the source nodes as the input data and the sink nodes as the final outputs. The basic operations in the workflow are well-defined, such as *buffer* and *clip*, and are provided in mature packages such as GeoPandas and QGIS tools. As a result, Level 2 autonomy, in which geoprocessing workflows are generated by LLMs, has already been demonstrated in studies such as LLM-Geo (Li & Ning, 2023).

Data preparation faces higher uncertainty. Questions arise, such as: What kind of data should be used? Where can it be found? Is the licensed data better than the publicly available data? How to consider the trade-off of the cost and expected result? Does it adequately cover the study area? Can the agent process large datasets? How does the quality of remote sensing imagery affect the project? What do acronyms in the attribute table mean? How should missing values and faulty data be handled? These data preparation challenges—spanning the domain's what, how, and why (Gahegan & Adams, 2014) —are numerous and complex. A Level 3 autonomous GIS is capable of addressing these uncertainties by preparing data with minimal human assistance. It can identify and access the most suitable data while making necessary compromises.

At Level 4, result-aware GIS evaluates results against expectations, identifying gaps and adjusting workflows, data sources, or models to close those gaps. Are the results satisfactory? What is the gap? How can the data sources and models be changed to narrow the gap? Additionally, the stochastic nature of the generative AI output may bring further uncertainties as the results from the same input vary among attempts from the same model (Akinboyewa et al., 2024; Atil et al., 2024), and the bias may increase due to training data and model architecture (Ferrara, 2023; Gallegos et al., 2024; Lin et al., 2024). The challenges at this level include the correct evaluation of the result and the prediction of improvement from the adjusted workflow or data sources. To address this, autonomous GIS agents must handle uncertainties in the data the associations between inputs, models, results, and vagueness more broadly.

Level 5, knowledge-aware autonomous GIS, is expected to run spatial analysis without human supervision and accumulate experience from practice. Like human analysts who learn from past attempts, summarize general rules, and apply their knowledge to similar tasks more efficiently, knowledge-aware GIS uses past experiences to better handle uncertainties in future tasks. As its knowledge and experience grow, task-related uncertainties will be gradually reduced by accumulated strategies, and the system will become increasingly autonomous, eventually reaching full autonomy. Note that full autonomy does not mean the elimination of communication between autonomous GIS and humans, as the system still needs to collect task context from users to reduce uncertainty, such as the budget or resource limitation for a sidewalk renovation project proposal, if the system cannot find such information in the task description. The knowledge-aware level emphasizes a specific pathway to achieving the self-growing goal—learning through past practice, experience, and other external documentation. We believe that iterative, rial-and-error learning processing is the foundation for advancing autonomous GIS to its highest potential.

## 2.4 Scales of autonomous GIS

To handle complex tasks, autonomous GIS must possess enough complexity and resources regarding *time*, *computing power*, *storage*, and *strategy* (Schmidgall et al., 2025). Here, *strategy* refers to the introduction of accessible resources and guidance on how to react to various scenarios based on given resources. Based on accessible resources, we categorize autonomous GIS into three scales: local, centralized, and infrastructure (Figure 4). These scales are defined by the resources available to the agent rather than the scale of the decision-core, which can be local, online, or hybrid across all scales.

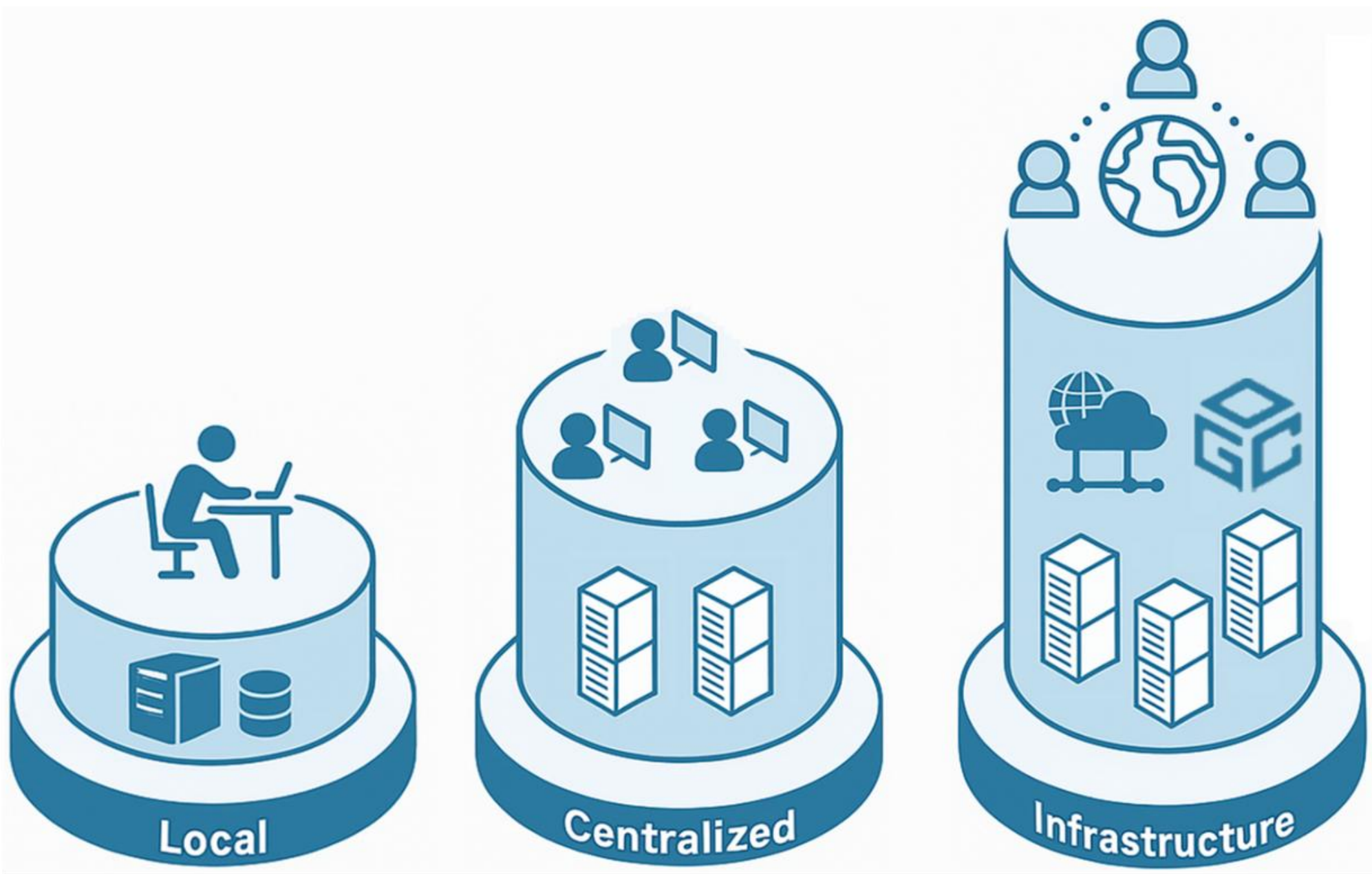

- Run on local computers
- Use computational resources of the computer
- Limited analysis strategies
- Single user

- Run on high-end servers or computing clusters
- Centralized management
- Powerful analysis strategies
- Multiple users

- Distributed agents among institutions
- Collaborations between centralized agents and local agents
- Inter-operational standards and protocols
- Public users

**Figure 4**. Scales of autonomous GIS

### 2.4.1 Local scale

At the local scale, an autonomous GIS implementation operates on a single computer, utilizing its computational resources, including CPU/GPU, memory, and storage. This scale is designed for individual users with limited resources and geoprocessing capabilities. The AI models for *decision-making* can either reside locally or be remotely connected, but the hardware capabilities of a single machine constrain the system's performance. For example, a local autonomous GIS can model the regional spread of infectious diseases using tabular electronic medical records and spatial data such as points of interest (POI) and boundaries within a reasonable time. However, its limited computational power makes it unsuitable for tackling large-scale or complex analyses. The GIS Copilot for QGIS (Akinboyewa et al., 2024) serves as a prototype of a Level 2 autonomous GIS operating at the local scale.

### 2.4.2 Centralized scale

At the centralized scale, the autonomous GIS leverages the computing power of a server cluster, enabling it to perform more advanced and data- and computing-intensive spatial analyses. It serves multiple users and is typically owned and maintained by one or several organizations. Centralized systems can process massive datasets and run sophisticated models, such as global climate simulations or country-level object detection from satellite images. By distributing workloads across the server cluster, centralized autonomous GIS could significantly expand the range and complexity of tasks compared to the local scale. Google Earth Engine (Gorelick et al., 2017) is an example of a geospatial running environment for centralized autonomous GIS. NASA Earth Copilot (Bryson, 2024) is one of the first attempts at implementing agents at this level.

### 2.4.3 Infrastructure scale

We envision the infrastructure scale as the most advanced level of autonomous GIS, built on geospatial cyberinfrastructure (Yang et al., 2010) with inter-operational standards and protocols. This scale integrates massive distributed computing resources and supports collaboration among institutions, enabling large-scale interdisciplinary research and public accessibility. At this scale, we expect the GIS agents from multiple institutions could work collaboratively to conduct large-scale spatial analysis, such as complex social or physical simulations that require extensive computing resources. It can also serve massive public users simultaneously for typical spatial analysis tasks, making geographic analysis and research more accessible to all. These systems rely on a robust ecosystem of geospatial algorithms, models, tools, and services hosted and maintained by various contributors ( Vandewalle et al., 2021; Wang, 2013). Infrastructure-scale autonomous GIS broadens access to geographic research as an extension of CyberGIS infrastructure (Michels et al., 2024) and national spatial data infrastructures (Masser, 1999). It allows governmental agencies, research communities, and individuals to contribute to and benefit from its capabilities. Meaningful questions and investigation results can be documented, shared, and reproduced, often with appropriate permissions, moving beyond traditional reliance on scientific journal publications.

Ideally, the more resources an agent can use, the more complex and challenging phenomena it can analyze. However, achieving autonomous goals becomes increasingly difficult as the scale expands because development and debugging costs rise accordingly, and there is no guarantee that diminishing returns put an end to growing a model's ability by adding resources. A key challenge lies in establishing secure standards, protocols and policies to let generative AI manage computational resources. At the current research stage of autonomous GIS, we prioritize the local scale, as it forms the foundation for higher scales. Commercial organizations are better positioned to explore centralized and infrastructure-level agents, whereas scholars can focus on developing standards and protocols for agents running on geospatial infrastructure.

## 3 Current Research Landscape Towards Autonomous GIS

As discussed, we view autonomous GIS as a new GIS paradigm that leverages the success of AI and existing GIS, to be implemented into various autonomous GIS agents. Autonomous GIS agents can be specialized to focus on a specific type of task, such as data collection or cartography, and these specialized agents can collaborate to accomplish complex tasks. In this session, we provide an overview of the current research landscape of autonomous GIS. Specifically, we review recent studies on autonomous GIS agents and highlight some notable examples. Other foundation technologies, such as LLM fine-tuning and geospatial foundation models, will be briefly mentioned in this paper.

Table 1 summarizes existing explorations, featuring 16 GIS agents (1–16), three remote sensing agents (17-19), four data science agents (20–23), and one research agent (24). Per literature, despite the immense potential, research and development of autonomous GIS agents are still in their early stages. Although there are attempts on general agents, such as Manus (Manus AI, 2025), they lack GIS knowledge and skills and may not be suitable for data-intensive spatial analysis tasks. Note that this table is not a comprehensive or exhaustive literature review for agents in the four mentioned fields. Instead, it aims to provide a rough research landscape in this rapidly developing area.

Most GIS agents adopted LLMs as the decision core to generate and implement workflows for spatial analysis, data retrieval, cartography, image analysis, and change detection. Because these agents can autonomously execute data processing workflows without human intervention, we label them autonomous Level 2. A few of them, such as NASA Earth Copilot (Bryson, 2024), were developed at a *Centralized* scale to serve the data request from a massive data inventory for multiple users. The remote sensing agents Guo et al., 2024; Liu et al., 2024) emphasize image understanding and operating abilities.

The data science agents reviewed here demonstrate greater autonomy when adopting the autonomous GIS levels. For example, AutoKaggle (Li et al., 2024) and DS-Agent (Guo et al., 2024) can clean data, which is *data-aware*, to some degree. Data Interpreter (Hong et al., 2024) and Data Science Agent (Google LLC, 2024) were able to adjust data processing workflows as needed by examining the results with a trial-and-error approach, so we categorize them as Level 4, *results-aware*. It is worth noting that the data science agents were designed to process tabular data, which is relatively straightforward for agents to understand and process, compared with the geospatial data which often comes in varied formats, modalities, and spatiotemporal resolutions. GIScience researchers have started fine-tuning vision LLMs to read maps (Zhang et al., 2025), the foundation for result interpretation of geospatial analysis.

Agent Laboratory (Schmidgall et al., 2025) demonstrates the potential for scientific research using autonomous agents. Acting as an assistant for AI researchers, this agent can conduct literature reviews, propose research questions, download data, create and debug code, and even write reports. This work provides a good example of an autonomous GIS capable of modeling tasks.

**Table 1**. Autonomous capacities of reviewed autonomous agents in GIS and broader data science.

| No. | Studies | Task Types | Autonomous functions | Autonomous goals | Autonomous level | Scale |
|---|---|---|---|---|---|---|
| 1 | LLM-Geo (Z. Li & Ning, 2023) | Spatial analysis | Decision-making, Data-operating | Self-generating, Self-executing | Level 2 | Local |
| 2 | GIS Copilot (Akinboyewa et al., 2024) | Spatial analysis | Decision-making, Data-operating | Self-generating, Self-executing | Level 2 | Local |
| 3 | IntelliGeo (Mahdi Farnaghi et al., 2024) | Spatial analysis | Decision-making, Data-operating | Self-generating, Self-executing | Level 2 | Local |
| 4 | ShapefileGPT ( Lin et al., 2024) | Spatial analysis | Decision-making, Data-operating | Self-generating, Self-executing | Level 2 | Local |
| 5 | GlobeFlowGPT (Kononykhin et al., 2024) | Spatial analysis | Decision-making, Data-preparation | Self-generating, Self-executing | Level 2 | Local |
| 6 | GeoAgent (Chen et al., 2024) | Spatial analysis | Decision-making, Data-operating | Self-generating, Self-executing | Level 2 | Local |
| 7 | GeoGPT (Zhang, Wei, et al., 2024) | Spatial analysis | Decision-making, Data-operating | Self-generating, Self-executing | Level 2 | Local |
| 8 | LLM-Find (Ning et al., 2025) | Geospatial data retrieval | Decision-making, Data-preparation | Self-generating, Self-executing | Level 2 | Local |
| 9 | ChatGeoAI (Mansourian & Oucheikh, 2024) | Geospatial data retrieval | Decision-making, Data-preparation, Data-operating | Self-generating, Self-executing | Level 2 | Local |
| 10 | Geode (Gupta et al., 2024) | Geospatial data retrieval and question-answering | Decision-making, Data-preparation | Self-generating, Self-executing | Level 2 | Local |
| 11 | NASA Earth Copilot (Bryson, 2024) | Geospatial data retrieval | Decision-making | Self-generating, Self-executing | Level 2 | Centralized |
| 12 | Geoforge (Ageospatial, 2024) | Geospatial data retrieval, spatial analysis, | Decision-making, Data-preparation, Data-operating | Self-generating, Self-executing | Level 2 | Centralized |
| 13 | MapGPT (Y. Zhang, He, et al., 2024) | Cartography | Decision making, Data-operating | Self-generating, Self-executing | Level 2 | Local |
| 14 | GeoMap Agent ( Huang et al., 2025) | Map understanding | Decision-making, Data-operating | Self-executing | Level 2 | Local |
| 15 | GeoAgent ( Huang et al., 2024) | Address Standardization | Decision-making | Self-generating, Self-executing | Level 2 | Local |
| 16 | GeoLLM-Engine* (Singh et al., 2024) | Geospatial Copilot Agent parallelization | - | - | Level 2 | Centralized |
| 17 | Geospatial Reasoning (Schottlander & Shekel, 2025) | GIS, Remote sensing image analysis and reasoning | Decision-making, Data-operating | Self-generating, Self-executing | Level 2 | Centralized |
| 18 | Remote Sensing ChatGPT ( Guo et al., 2024) | Remote sensing image analysis | Decision-making, Data-operating | Self-generating, Self-executing | Level 2 | Local |
| 19 | Change-Agent ( Liu et al., 2024) | Remote sensing image change detection | Decision-making, Data-operating | Self-generating, Self-executing | Level 2 | Local |
| 20 | AutoKaggle ( Li et al., 2024) | Data science agent | Decision-making, Data-operating | Self-generating, Self-executing | Level 3 | Local |
| 21 | DS-Agent ( Guo et al., 2024) | Data science agent based on cases | Decision-making, Data-operating | Self-generating, Self-executing | Level 3 | Local |
| 22 | Data Interpreter ( Hong et al., 2024) | Data science agent | Decision-making, Data-operating | Self-generating, Self-executing | Level 4 | Local |
| 23 | Data Science Agent** (Google LLC, 2024) | Data science agent | Decision-making, Data-operating | Self-generating, Self-executing | Level 4 | Centralized |
| 24 | Agent Laboratory (Schmidgall et al., 2025) | Assistant for AI Research | Decision-making, Data-preparation, Data-operating | Self-generating, Self-executing, Self-organizing | Level 4 | Local |

*GeoLLM-Engine* (Singh et al., 2024) is a framework to scale up agents, such as replicating an agent into hundreds of copies thatcan run simultaneously to help address tasks.

**Data Science Agent (Google LLC, 2024) is an online service allowing users to upload data; it is for a single user, but we think such an elastic manner can be viewed as centralized, as the provider — although not currently, but rather theoretically — can easily provide additional computing power to users upon request.

## 4 Case Studies

This section presents four case studies to illustrate how autonomous GIS automates data retrieval, spatial analysis, and cartography involving vector and raster data. The first case examines LLM-Find, a geospatial data retrieval agent that automates data acquisition based on natural language inputs. The second case highlights LLM-Geo, a spatial analysis agent that independently executes geoprocessing workflows to assess school walkability. The third case introduces LLM-Cat, a proof-of-concept cartography agent, to demonstrate the possibility of vision-based autonomous cartography. The final case explores a different type of autonomous GIS agent: GIS Copilot, which assists users in performing terrain analysis using well-established GIS tools. All these agents are developed for local machines at Level 2 (Workflow-aware), with the functionality of *decision-making* and *data operation*, except LLM-Cat, which focuses on *data-preparation*.

### 4.1 Autonomous GIS agent for geospatial data retrieval

Geospatial data retrieval is critical in enabling effective scientific analysis and increasing the accessibility and usability of important data ( Li, 2018). Automating this process will significantly reduce data preparation time and accelerate analysis. LLM-Find is a geospatial data retrieval agent (Ning et al., 2025) that human users or other GIS agents can use to request data via natural language. The agent is implemented in two formats: a QGIS plugin that allows users to directly manipulate the downloaded data in QGIS and a Python module that offers more flexibility when integrated into other systems. The agent contains 7 data sources by default, such as OpenStreetMap, Esri World Image basemap, and global digital elevation models (DEM). It also supports customized data sources using a plug-and-play manner. In this case, we use LLM-Find to download geospatial data for school walkability analysis in Columbia, South Carolina, USA. The target data includes vector data from OpenStreetMap such as sidewalk polylines, road polylines, and school points, as well as high-resolution remote sensing images from Esri basemap.

We input the data requests individually (Table 2) to the LLM-Find (QGIS plugin version). As shown in Figure 5, it successfully selected and downloaded the correct data sources and requested data within several minutes, indicating that it is possible to fetch data automatically based on natural language commands for spatial analysis tasks in autonomous GIS.

**Table 2.** User input data requests for school walkability assessment

| No. | Task Input |
|---|---|
| **1** | Download sidewalk data in Columbia, South Carolina, USA. |
| **2** | Download the road network (excluding footways and service ways) data in Columbia, South Carolina, USA. |
| **3** | Download all schools in Columbia, South Carolina, USA, and convert them to points. |
| **4** | Download images for Columbia, South Carolina, USA, at level 15. |

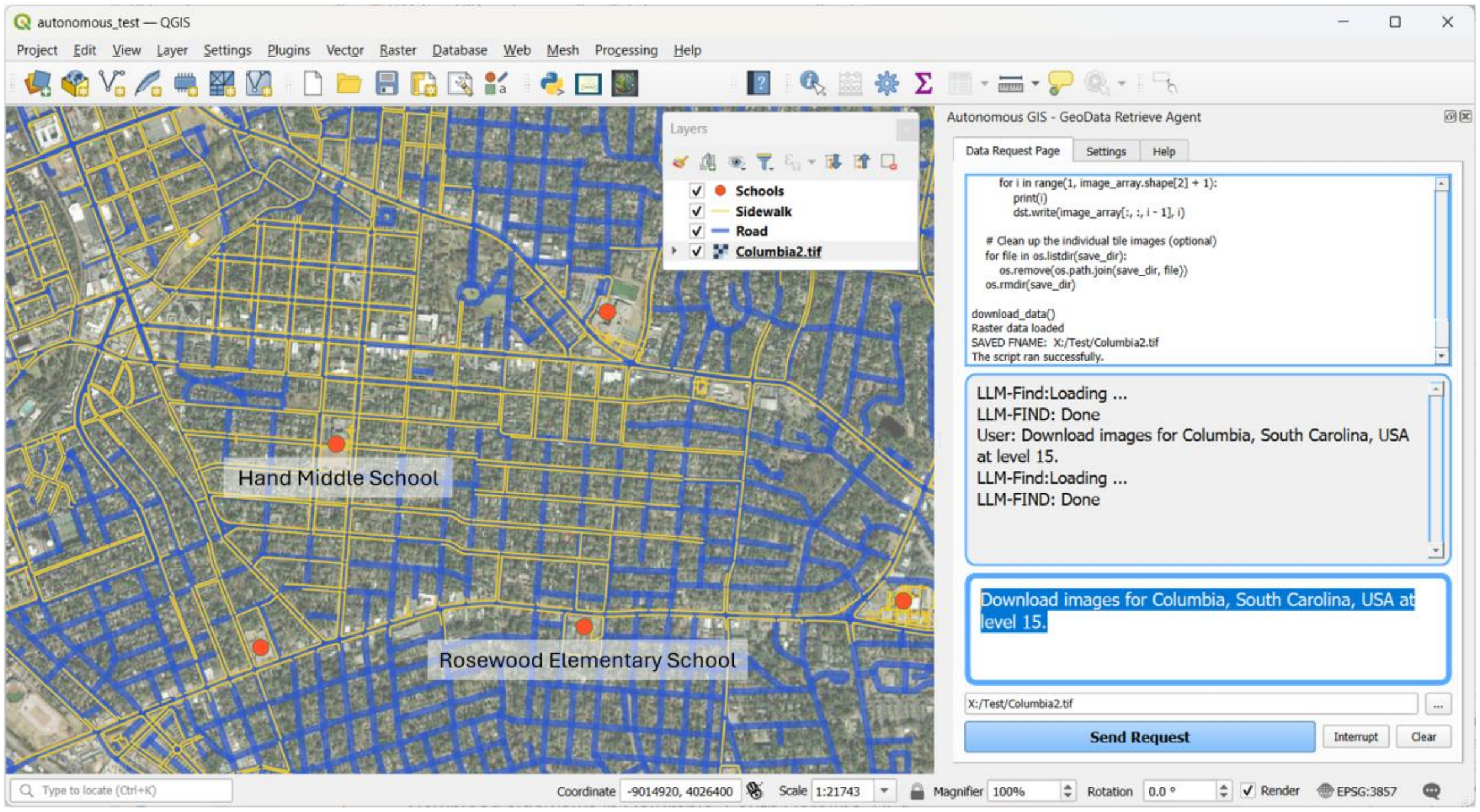

**Figure 5**. Using a geospatial data retrieval agent to download various geospatial data, including school location, sidewalks, road network, and high-resolution remote sensing images from OpenStreetMap and Esri World Image. Note that we manually change the data symbology for a clear visualization. LLM-Find's source code and over 70 data request examples can be found on *GitHub*. The QGIS plugin can be downloaded from the official QGIS plugin *website*.

## 4.2 Autonomous GIS agent for spatial analysis

LLM-Geo is the first systematically documented GIS agent designed to perform spatial analysis autonomously by receiving requests in natural languages (Li & Ning, 2023). In this case, we used LLM-Geo to conduct a spatial analysis task: assessing the school walkability based on the vector data downloaded by the data retrieval agent, LLM-Find (see section 4.1). Figure 6 illustrates how LLM-Geo works for answering spatial questions. The process starts with user input, including the question and relevant data sources such as online URLs, data services, and API documentation. LLM-Geo then decomposes the spatial problem into multiple steps and connects those steps as a directed acyclic graph to form a geoprocessing workflow. Each operation node is implemented as an executable Python function, which is then assembled into a complete program to generate the final answer, such as maps, charts, reports, or new datasets.

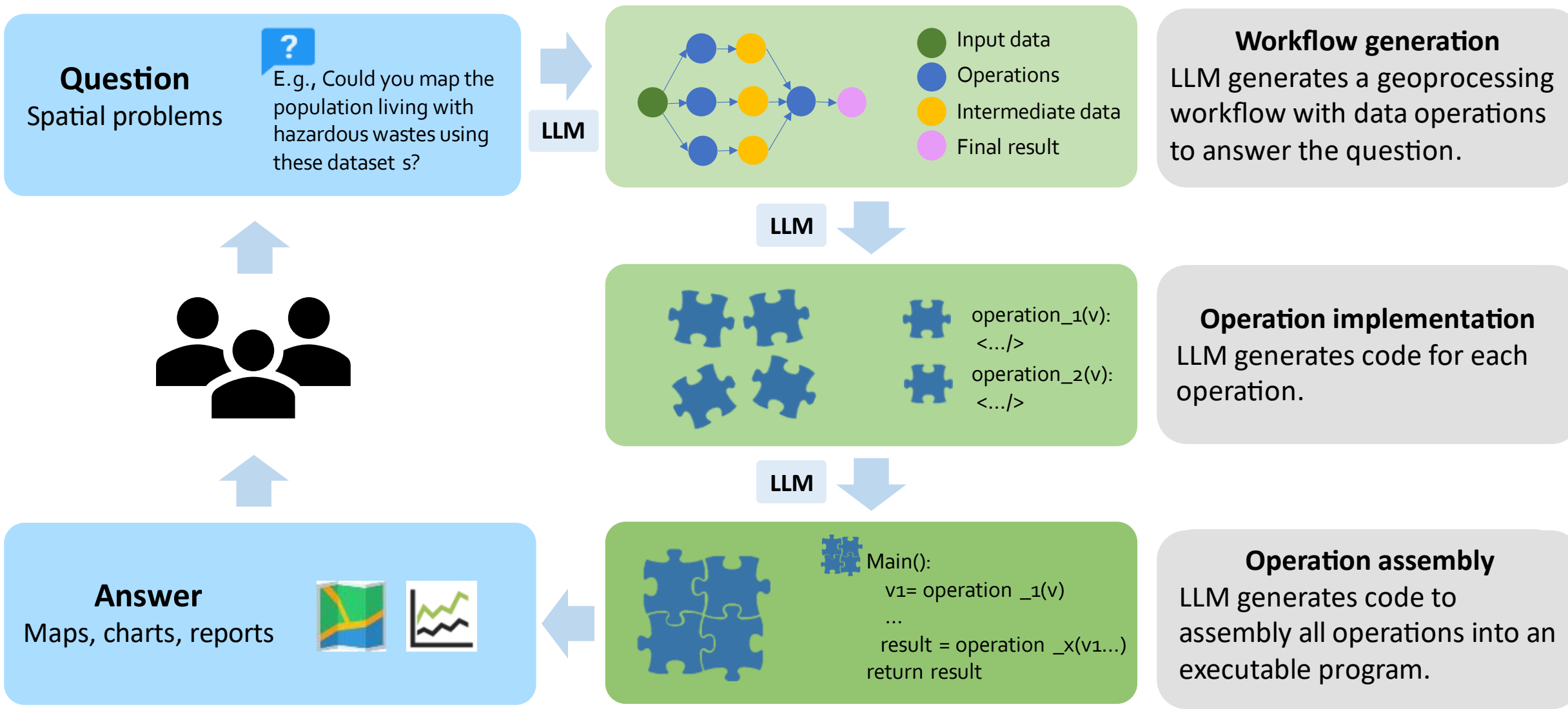

**Figure 6.** Illustration of how LLM-Geo works: from user query to geoprocessing workflow, executable code, and final output. (Adapted from Li and Ning, 2023)

We asked LLM-Geo to compute the school walkability score using the provided data within a 1 km buffer; the score is defined as the ratio of sidewalk length to the road length. After obtaining the score, LLM-Geo must draw a map for each school, showing the school name, the associated score, sidewalks, buffering circle, and the OpenStreetMap basemap. Although this task is relatively straightforward for human GIS analysts, it is laborious and time-consuming. We input the requirements and data location to LLM-Geo as shown in Box 1, and it accomplished the task within 5 minutes during our test (Figure 7). Although sometimes the expected results were not returned in the first run, LLM-Geo could self-debug the code issues and return the correct results in the second attempt autonomously. Note that we do not specify which column of the school layer stores the school name; instead, a data understanding module is equipped to collect layer metadata, including column information, to feed into the decision-core.

**Box 1.** Input to LLM-Geo for school walkability assessment. The input contains two parts: requirements and data locations, both in natural language.

---

TASK = You need to assess the walkability of all schools in Columbia City. Requirements:
1) Using 1 km buffer zone as the study area for each school, compute each school's walkability score, which is the ratio of the sidewalk length to the road length in the buffer zone.
2) Draw a map for each school, using the school name and the walkability score as the map title, while showing the sidewalks and the buffer zone on an OpenStreetMap basemap.

DATA_LOCATIONS = [
1. Columbia school (points): https://github.com/gladcolor/Columbia_schools.gpkg.
2. Columbia road network (polyline): https://github.com/gladcolor/Columbia_road.gpkg.
3. Columbia sidewalk network (polyline): https://github.com/gladcolor/Columbia_sidewalks.gpkg. ]

---

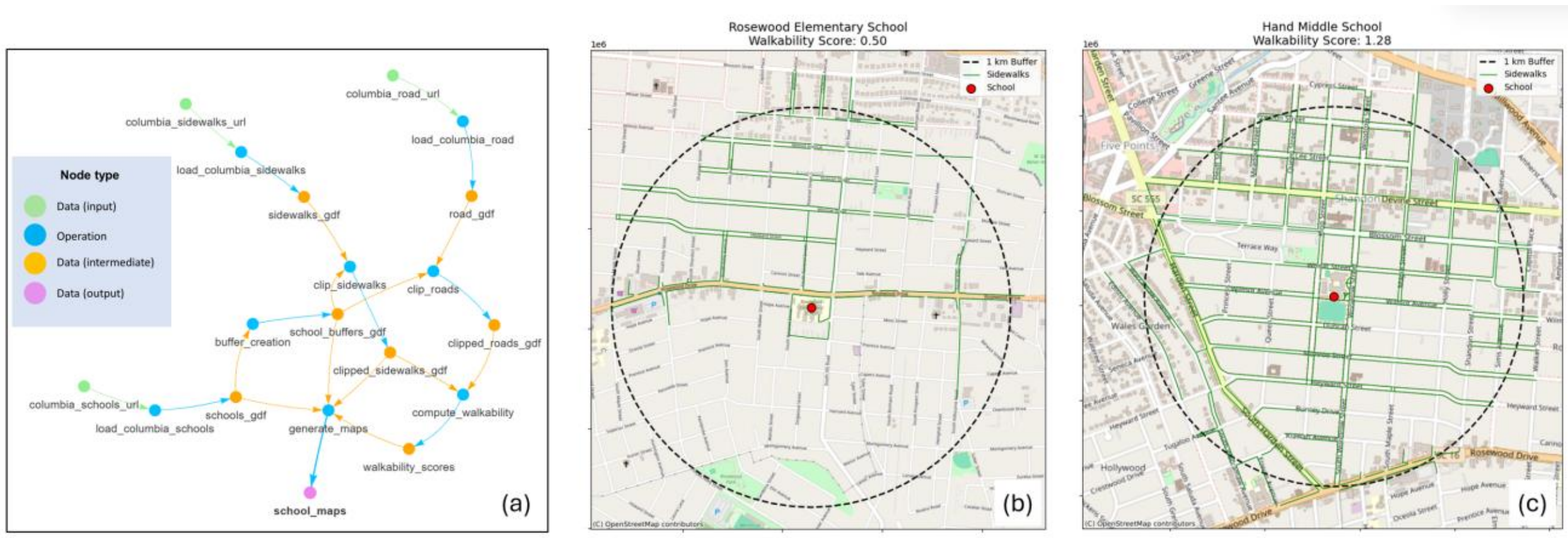

**Figure 7**. The results of the school walkability assessment by LLM-Geo. (a) The generated workflow. (b), (c) Two school map examples that show the walkability score, buffer, and sidewalks. The source code of LLM-Geo and more case studies can be found on *GitHub*.

### 4.3 Autonomous GIS agent for cartography

We prototyped a cartography agent named LLM-Cat to make simple maps. It utilized the vision capability of GPT-4o to demonstrate the possibility of vision-based autonomous cartography. LLM-Cat accepted map-making requests in natural language and generated maps using Python code. Because cartography is a vision-based and iterative process, it requires map cognition and understanding the causation between the code (or GIS operations) and the map. Trained cartographers regularly make multiple attempts and modifications before the map fits the context and requirements. We argue that generative AI with the vision model can make maps by mimicking the cartographers' behaviors; therefore, LLM-Cat is designed to work iteratively: 1) it generates an initial map based on the given data and requirements; 2) it reviews the map and points out one issue, such as the legend is missing; 3) it revises the code to solve the issue and generates a new map; and 4) it repeats step 2) and 3) until it reaches the maximum iteration or it thinks there are no more issues.

Box 2 shows the input of a use case of LLM-Cat; it requires the agent to create a map showing the hospital locations in Pennsylvania, USA. The map requirements are detailed in natural language, such as title, north arrow, and basemap. Data locations are also provided. Figure 8 shows the maps generated from 4 rounds (Round 0, 3, 7, and 10). After 10 rounds, LLM-Cat achieved multiple improvements: 1) increasing the title fonts; 2) re-arranging the "Designer" annotation; 3) changing the height of the north arrow; 4) replacing the hospital point color; 5) turning up the transparency of the basemap. These improvements make the map look better than Round 0 (initial map). We noticed that the backend AI model (GPT-4o) has relatively weak map aesthetics and cartography skills, so the final map does not look very appealing although it does meet all the requirements. This is among the first attempts among others (Memduhoğlu, 2025; Xu & Tao, 2024; Zhang et al., 2024) to use AI for autonomous cartography, and the result is promising. Future research can incorporate the target audience's background, for example, asking the agent to make maps for children, color-blind viewers, or a specific domain, such as the health science community.

**Box 2**. Input to LLM-Cat for a map-making task; it is asked to create a map to show the hospital locations in Pennsylvania, USA. The input contains two parts: task requirements and data locations, both in natural language.

TASK = "1. Create a map showing the hospital locations in Pennsylvania.
2. Carefully design the map and make it beautiful and aesthetically appealing.
3. Use a local map projection of metric units.
4. The map needs a title, a north arrow, a scale bar, a legend, and a designer. The designer is 'LMM-Cat'.
5. The map dimension is the landscape letter (11*8 inches) size, and the DPI is 100.
6. An overview submap in the main map is needed to show Pennsylvania's location in the USA.
7. Add an OpenStreetMap basemap.
Note that all given data are in GeoPackage format, and each .gpkg file is limited to one layer; there is no need to load it using a layer name. All data are projected toEPSG4326."

DATA_LOCATIONS = "1. Hospital point data: D:\PA_hospital.gpkg.
2. Pennsylvania county boundary polygon data: D:\PA_County_boundary.gpkg.
3. Pennsylvania state boundary polygon data: D:\PA_State_boundary.gpkg.
4. Contiguous USA boundary polygon data: D:\USA_boundary.gpkg."

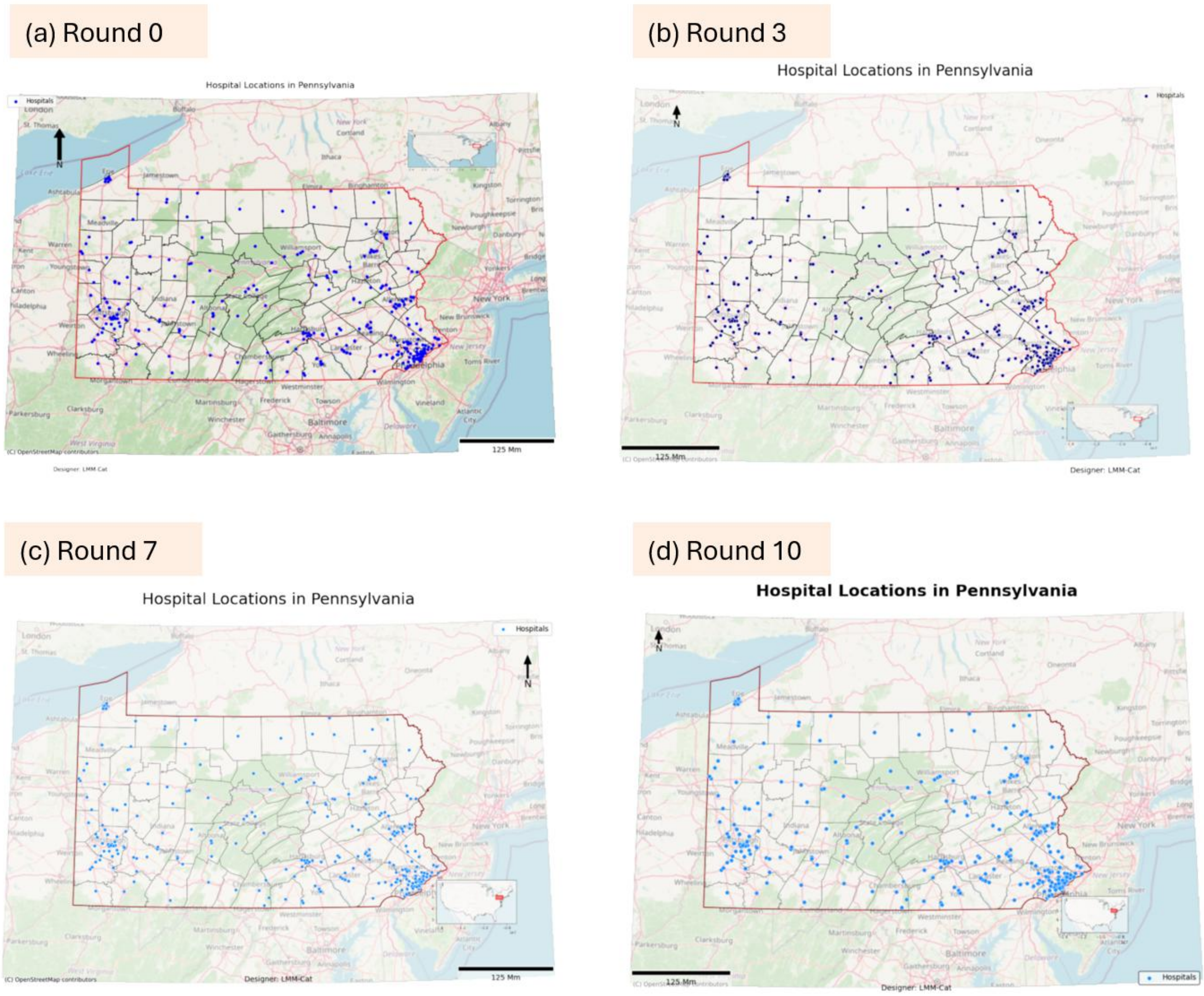

**Figure 8**. The maps from multiple rounds (Round 0, 3, 7, and 10) of the cartography agent LLM-Cat. Note that the scale bar unit should be "km". The backed GPT-4o model has noticed this error and set the unit parameter to "km" as the document of the Python scale bar package (*matplotlib_scalebar*), but the shown unit is "Mm" which may be caused by *matplotlib_scalebar* package bugs. The source code of LLM-Cat and more case studies can be found on [GitHub](#).

### 4.4 Autonomous GIS agent as GIS Copilot

In this section, we demonstrate a use case of the GIS Copilot (Akinboyewa et al., 2024) as another form of autonomous GIS agent. This copilot is implemented as a QGIS plugin, aiming to assist QGIS users in using the mature GIS analysis tools in QGIS, rather than redeveloping them on the fly by the spatial analysis agent such as LLM-Geo (Li & Ning, 2023), which is often error-prone. This case shows how autonomous GIS uses the *Slope*, *Aspect*, and *Terrain Ruggedness Index* tools in QGIS. Before conducting these tasks, the GIS Copilot performed basic data operations such as raster merging, reprojecting, and clipping as the study spans the adjacent regions of four DEM tiles. Box 3 lists the text input to the Copilot. The request does not explicitly mention which tools should be used for this task but notes the reprojecting operation for the Copilot because the backend GPT-4o model has relatively little knowledge in coordinate reference systems when manipulating geospatial data. Figure 9 displays the results of the request. The Copilot accomplished all tasks in one attempt within 5 minutes; it reprojected the data, merged the four DEM datasets, chipped the data based on the county boundary, and finally generated the slope, aspect, and terrain ruggedness index without human intervention.

**Box 3.** Input to LLM-Geo for school walkability assessment. The input contains requirements and data locations, both in natural language.

> Merge the four DEMs into a single raster and perform terrain characteristic analysis for Richland County, South Carolina, USA, including slope, aspect, and terrain ruggedness index (TRI). Note that the given data are in EPSG 4326 (unit: degree), so you must reproject the data into EPSG 6569 (unit: meter) before the terrain analysis.

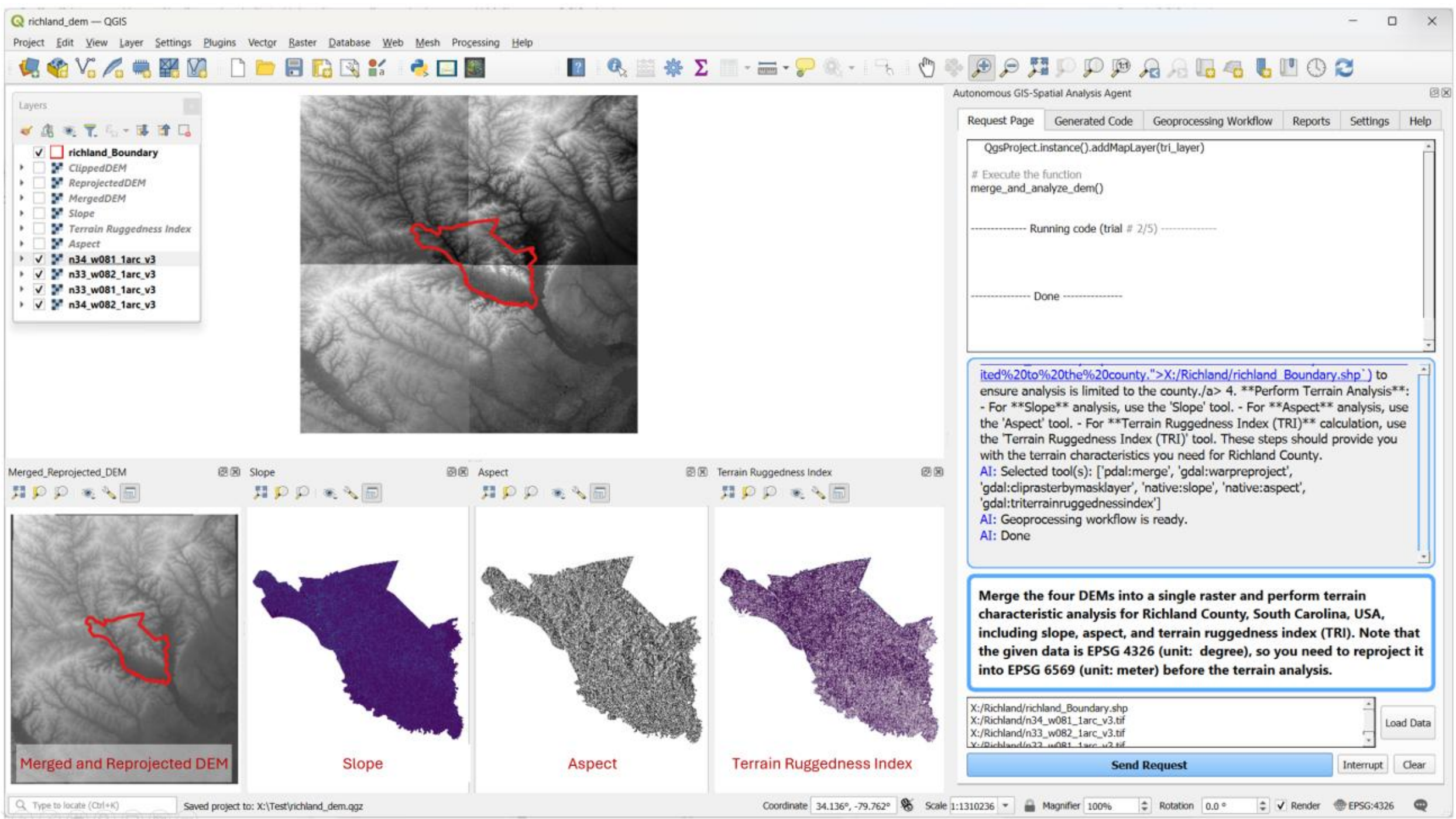

**Figure 9**. The results of the DEM operations by the GIS Copilot. The GIS Copilot can be downloaded from the official QGIS plugin *website*, and the source code and more case studies are available on *GitHub*.

## 5 Challenges and Research Agenda

In this section, we outline the challenges of advancing autonomous GIS and propose a research agenda for this emerging field. We envision a new generation of GIS powered by AI, capable of performing spatial analyses with and beyond the expertise of a human analyst. The development of such a GeoMachina which was suggested as a community-driven Moonshot by Janowicz et al. (2020) —"Can we develop an artificial GIS analyst that passes a domain-specific Turing Test by 2030?"—is poised to become a reality. To achieve this vision, we prioritize several key research areas focused on autonomous GIS at the local scale. Whereas commercial organizations may explore centralized agent systems, scholars and practitioners can advance the theoretical knowledge and lead the development of standards and protocols to support autonomous GIS agents operating within the geospatial infrastructure.

## 5.1 Geospatial skill and knowledge probe for generative AI models

### 5.1.1 LLMs lack GIS-specific knowledge and skill

Generative AI models such as LLMs are mostly trained on publicly available online data. However, not all domain-specific knowledge is explicitly documented online, nor might they be sufficient to capture the complex interactions between humans and the environment. Consequently, developing accurate prompts or instructions for LLMs often involves a time-consuming trial-and-error process. This issue is exacerbated by the lack of transparency regarding generative AI's capabilities in geospatial analysis, often caused by the failure to learn concepts of location and space automatically. For instance, GPT-4 shows misunderstandings in critical GIS concepts, such as incorrectly computing distances and areas using geographic coordinate systems instead of projected systems, or falling short of understanding topology (Majic et al., 2024). Similarly, it misinterprets some OpenStreetMap Overpass API statements, making it difficult to correct its behavior even with carefully crafted prompts (Akinboyewa et al., 2024; Li & Ning, 2023; Ning et al., 2025).

AI's lack of GIS-specific skills significantly limits its capabilities as the decision-core ("brain") of autonomous GIS. Given the expectation of data operation and analysis for autonomous GIS, we also emphasize skills—practical abilities to operate data and solve problems. For example, the skill of data reprojection involves using GIS tools or algorithms to transform data into a different coordinate system at the level of precision/accuracy desired. AI can output a text description of the steps and knowledge involved in reprojection; however, a plausible description does not guarantee the success of the data operation and the correctness of the result. Making an analogy, a student passing the driving knowledge test does not guarantee that the student can drive a car safely.

### 5.1.2 Geospatial knowledge probe

From a task-agnostic viewpoint, generative AI models face limitations due to outdated training data and may lack new knowledge or information. For instance, AI may not reflect newly released algorithms or Python packages. Suppose an efficient practice of using an existing GIS package (for example, GeoPandas) to extract vector layer intersections is released (Nestler, 2024). In that case, AI cannot evaluate or adopt it without external intervention, so the autonomous GIS agent it powers may take unreasonable time to process the data. Although AI models can use an Internet search and web crawling tools to collect updated information, the timing of triggering these tools requires an extra mechanism. Another limitation of AI is the lack of domain-specific knowledge. Tasks such as identifying mariculture sites involve legal, geographic, and bureaucratic constraints that require knowledge beyond public Internet resources. Furthermore, this knowledge is so intrinsically location-specific, that it makes transferring skills across locations a very difficult challenge and limits the applicability of these automatically learned solutions. This highlights the need for AI to learn from external documents to expand the domain knowledge base, especially to avoid repeating mistakes in similar tasks. Retrieval-augmented generation (RAG) technology is commonly used to incorporate external information into the generative AI outputs. Fine-tuning mechanisms such as LoRA (Hu et al., 2021) are developed to enable AI to adapt and grow autonomously. Thus, identifying the missing AI knowledge becomes a prerequisite before fine-tuning.

A potential practice to probe the missing knowledge is comparing internal knowledge with external documents. Specifically, given a document, we can use LLM to create a title and an abstract (context only, without leaking critical content), and then direct the AI *agent* to write a replicated document according to the title and abstract. Next, we can use an LLM to compare the replicated document with the given document and then conclude whether AI has knowledge of the external document. For example, given a blog about the rapid implementation for obtaining the intersection of two massive vector layers in Python, we can ask LLM to write a post and a program about the same topic. If the LLM cannot replicate the content in the blog, it should store the blog, label it as valuable external knowledge, and use the core-updating function to learn the best practice provided in the blog. Another example of the knowledge probe is information updating. For instance, we can ask AI about the latest version of ArcGIS, and let the LLM compare the searched page. Such internal knowledge probes are needed to achieve the autonomous goal of self-growing.

Users may need to provide the AI with the required domain knowledge, context, and even governmental regulations regarding a specific task. This “human-in-the-loop” approach is practical but may require rounds of human interaction or intervention, lowering the autonomy. Detecting the missing knowledge and obtaining it later autonomously can reduce human interaction.

#### 5.1.3 Geospatial skill probe

The misalignment between AI’s kills and expected behaviors significantly reduces efficiency in developing autonomous GIS. Considering the skill sufficiency of a GIS analyst when facing a task, it can be roughly classified into three levels: 1) Full skill: analysts possess sufficient skills and confidence, knowing exactly what to do in tasks such as downloading all scanned historical geological maps, 2) Partial skill: analysts lack full confidence in accomplishing the task, for example, extracting geological information from the scanned maps, but they will test some possible solutions, and 3) No skill: analysts have no experience in the task, such as predicting critical mineral deposits from a geological database, so they turn to external resources such as Google to help. Similarly, for the autonomous GIS, knowing what “I” can do is the first step before planning geoprocessing workflows. In other words, it should be self-aware, knowing what it can and cannot do. This raises a critical question: Can AI evaluate its skills or capabilities in a specific domain?

To address this challenge, we believe a benchmark for basic skill testing is needed to evaluate the generative AI’s abilities in spatial analysis. Such a benchmark could include tasks like joining attributes with different data types such as float, string, and null, calculating a walkability index based on the ratio of sidewalk length to road length, or modeling the spread of infectious diseases using human mobility data. Akinboyewa et al. (2024) introduced a GIS Copilot benchmark with 110 test cases across three difficulty levels, marking the first attempt at autonomous GIS benchmarking. However, automated testing and comprehensive benchmarking remain underexplored.

## 5.2 Self-growing

In autonomous GIS, generative AI models are used as the decision-core to plan, implement, and execute solutions for given tasks. We expect autonomous GIS and its decision core to achieve the goal of self-growing: it should learn experiences from tasks and external documents for higher productivity and capability, like humans. In addition, autonomous GIS should be able to extend its capability if the user or organization's mission shifts. Technologically, it can be a fine-tuning process for the pre-trained decision-core using specific domain data. When fine-tuning the AI model that powered the autonomous GIS, the domain data mainly comes from the given tasks, generated code, and external documents. Fine-tuning changes a few weights of the model or the additional matrix, for example, a bypass matrix in LoRA (Hu et al., 2021), aligning the output distribution to the fine-tuning dataset. Autonomous GIS should have a module to enable self-fine-tuning. Interestingly, most current machine-learning-based AI systems are not based on continuous approaches.

### 5.2.1 Incremental and discrete self-growing

The decision core is the "brain" of autonomous GIS. It is not necessarily a single model. It can be a large foundation model or a combination of multiple large online foundation models and small local models. Each model may be skilled at specific tasks, such as modeling, workflow generation, programming, visual assessment, and writing reports. Therefore, each round of self-growing can be divided into small pieces with different skills. Practical methods are in two paths: 1) dividing decision core into separated experts and fine-tuning them using specific domain documents, and 2) hybrid methods combining prompt engineering, fine-tuning, and RAG. For example, the generated runnable code with verified results can be used as training data to fine-tune the model. The GIScience community has attempted to fine-tune LLMs for tool selection (Wei et al., 2024), spatial analysis (Mansourian & Oucheikh, 2024), and geospatial code (S. Hou et al., 2024). Scholars can further investigate autonomous fine-tuning for the decision core for specialized agents such as a data preparation agent. Since the number of trainable parameters is limited, the new knowledge or skills learned are also limited. Thus, fine-tuning multiple specialized agents to accomplish tasks collaboratively may be a practical method. Self-growing does not mean the growth of a single decision-core, but a set of them, and not every decision-core updates simultaneously. They update themselves discretely.

### 5.2.2 Enhancement of spatial programming skills of decision-core

Some studies revealed that state-of-the-art LLMs still have relatively weak programming skills in spatial analysis (Li & Ning, 2023; Ning et al., 2025). For example, they may ignore the differences in the coordinate systems and attribute data types among vector datasets when generating data processing code. Typically, scientists intensively test the various cases to probe missing knowledge and then explicitly incorporate that knowledge into the prompt for future tasks. We have broadly discussed the knowledge probe in Section 5.1; in this research question, the objective is to design a self-fine-tuning mechanism to incorporate the missing knowledge into the decision-core. Many issues need to be explored. For example, how many training samples are needed? Or is merely a checklist good enough? Note that checklists may represent multiple perspectives and application domains beyond geospatial data processing. For example, when asking a GIS agent to use a less well-known or newly released public health dataset for spatial analysis, the backend AI may have no information about the specific dataset. Thus, a checklist for public health analysis using AI is required. Such a checklist may be a long or brief document with concise general rules without training samples, requiring regular update. New technologies and knowledge will be generated continuously, and the AI models' internal knowledge will eventually become outdated; if they cannot catch up and incorporated the new knowledge, they may rely on a “search engine”, lowering the decision-making efficiency and bring extra uncertainty from external information.

#### 5.2.3 Continuous learning in self-growing

Self-growing enables autonomous GIS to incrementally enhance its capabilities through continuous learning. This goal assumes that not all past patterns will persist in the future. This assumption embraces dynamic domains like disaster response or public health, where data distributions and task requirements can shift unpredictably (Zheng et al., 2025). The missions of users and organizations will also change over time. These shifts, referred to as *concept drift* in machine learning, pose serious challenges for models that do not monitor or adapt to temporal changes (Gupta & Gahegan, 2020; Lu et al., 2018). We believe self-growing systems must evolve beyond static or discrete learning cycles and incorporate mechanisms for temporal monitoring and adaptive learning (Wu et al., 2024). Unlike traditional fine-tuning, which is typically triggered manually or periodically, these mechanisms would operate continuously or semi-continuously, detecting deviations in data characteristics and updating relevant components of the decision-core accordingly. This could involve lightweight fine-tuning on recent data, dynamic prompt revision based on temporal context, or periodic revalidation of performance benchmarks. Rather than treating time as an implicit dimension, we suggest that future autonomous GIS should explicitly model temporal variability as part of their growth strategy. This may include designing systems that can distinguish between stable and volatile knowledge domains, prioritize updates in areas experiencing rapid change, and allocate resources to refresh model accordingly (Peng et al., 2023).

### 5.3 Getting ready for autonomous geographic modeling

In this paper, geographic modeling refers to creating a logical series of steps and mathematical representation of the processes to predict (and hopefully, explain) the quantitative causality or association between geographic entities. Modeling is critical in geospatial analysis, aiming to uncover patterns and dynamics in societal and environmental contexts. Modeling addresses “why”, “how”, and “what if” questions, such as explaining biodiversity decline and predicting the process in areas with specific land-use changes. Models can also predict and optimize, exemplified by weather forecasting and route planning for package delivery. GIScience essentially converts a geographically embedded physical/human process and data into a workflow and computational solution.

We emphasize autonomous modeling because it can maximize the efficiency of searching for explanations or predictions while removing the technical barriers for non-GIS users. The autonomous GIS can run multiple models in parallel and refine them iteratively. An ensemble of models might be necessary when alternative theories are preferred to be included (e.g., as in weather forecasting). We expect this cost-effective and exhaustive model and parameter search to foster reasonable results without subjecting it to human analysts’ expertise and working time limitations.

When conducting geographic modeling, human GIS analysts tend to iteratively pick up familiar models, collect needed data, set up model parameters, run the model, and assess the results. Such a process is demanding and time-consuming, even for a simple linear regression model. It involves considerable perseverance, domain knowledge, theoretical framing, and reasoning. Generative AI also seems to have such abilities, although still in the preliminary stage. They can visualize the geospatial data and results, generate code on the fly, and assess the result. Moreover, they can run overtime without stopping the search for optimized models and parameters once sufficient computational resources are secured. Thus, we discuss the possibility of autonomous modeling in this session.

#### 5.3.1 Can autonomous GIS autonomously model geographic phenomena?

The implementation of autonomous GIS provides the technological pathway toward autonomous modeling of geographic phenomena - in other words, it makes the autonomization of geographic science possible. Generally, the key of autonomous sciences is to transform model and parameter search into a process of optimization, ideally solvable within reasonable time and resource constraints. This often involves framing the optimization task as a problem in the complexity class NP (nondeterministic polynomial time) (Fortnow, 2009). If autonomous agents can propose and verify candidate solutions in a cost-effective manner—for example, through rapid computation or experimentation—the problem may be tractable . Gahegan (2020) conceptually proposed autonomous geographic modeling. He expected a system that could discover explainable models in a data-driven and causal referencing manner. Such systems construct inductive process graph models from a “soup” of theory fragments, rules, constraints, and observations, and then search for the best model that fits the observations. Such a symbolic regression (Makke & Chawla, 2024) method can be used in autonomous modeling, whereas the AI can initialize and adjust the model parameters. However, symbolic regression is not a universal modeling method for all geographical phenomena, so the “model base” needs to contain other models.

Various algorithms and models can be used in geospatial analysis, such as statistical and regression-based models (e.g., Geographically Weighted Regression), cluster and classification (K-means, SaTScan), machine learning and AI models (e.g., Support Vector Machine, Random Forest, Neural Network), spatial interaction models (gravity model), physical models (hydrological models), simulation models (cellular automata, agent-based models), raster-based models (terrain analysis, interpolation), network (shortest path algorithms), time serial, spatial optimization (multi-criteria decision analysis, genetic algorithms), causality modeling, and many other recently developed models such as Spatial Variability Aware Deep Neural Networks ( Gupta et al., 2021), Significant DBSCAN+ (Xie et al., 2021), just to name a few. The geoprocessing workflow in spatial analysis implements the selected applicable model. Hence, can autonomous GIS use these algorithms and models appropriately for various phenomena? This is challenging because algorithms and models require optimal parameters or iterative implementation and adjustment. For example, what variables should be used in a regression model? How to create an agent-based model for a susceptible-exposed-infected-recovered (SEIR) epidemic study (Kaddar et al., 2011)?

The scientific community has investigated auto modeling for a long time, especially in automatic machine learning (AutoML) (Salehin et al., 2024), such as tree-based pipeline optimization tool (TPOT) (Olson & Moore, 2016), Auto-Sklearn (Feurer et al., 2022) and AutoGluon (AutoGluon, 2019/2025). In the agent domain, Google (Google LLC, 2024) has developed a data science agent to conduct data cleaning, data analyzing exploration, modeling, and model evaluation. Like LLM-Geo (Li & Ning, 2023), Data Interpreter (S. Hong et al., 2024) is an agent for data science using the directed acyclic graph to organize data modeling and analysis sub-tasks. The authors also proposed a hierarchical task graph and graph refinement, similar to human analysts' divide-and-conquer and trial-and-error strategies for problem-solving. Guo et al. (2024) developed a data science agent based on the Kaggle case; it retrieves similar cases in Kaggle for the given task and revises the code to model the new task. Regarding model selection, many practices (e.g., Akinboyewa et al., 2024; Ning et al., 2025) adopt a straightforward – push a list of candidate resources (models, tools, and data) and descriptions to the backend LLM and then expect the LLM select an appropriate resource for the task.

We encourage the GIScience community to explore effective methods for model selection, such as fine-tuning a small LLM to suggest suitable models. One of the current urgent missions is to prepare a "case base" by collecting articles using geospatial models; this case base contains human insights into modeling phenomena from the geographic perspective.

#### 5.3.2 Meta-models: models for modeling

As autonomous GIS is expected to select a suitable modeling method for a geographic phenomenon, we may ask: what are the guidelines? Should we rely purely on AI's selection? Or can we impose a thinking framework for AI to better model the world? Autonomous GIS is supposed to answer "why", "how", and "what if" questions via geospatial data modeling. What kind of questions are autonomous GIS able to answer? For example, why does HIV prevalence vary among US counties (Li et al., 2021)? Should autonomous GIS model the phenomena via the perspective of geographic disparities? Or should it model the phenomena in a geographic space? What if access to HIV prevention services were improved in rural counties, how might that impact future prevalence patterns?

From a broad perspective, a pressing question is the temporal and spatial frame consideration (Goodchild & Li, 2021; Kwan, 2012) when autonomous GIS conducts analyses. For example, should autonomous GIS base its predictions on recent climate trends from the past decade in flood risk assessment, or should it incorporate long-term hydrological patterns spanning the last century? How do we teach AI heuristic frameworks or thinking strategies, such as systems science (Shalizi, 2006) or dialectics (Goonewardena, 2023)? In pursuing answers, our ultimate goal is to advance from artificial intelligence toward a more human-centered form of intelligence in data science—one that prioritizes moral discernment, empathy, and long-term, holistic judgment.

There are many heuristic frameworks and thinking strategies for scientific modeling, in other words, meta-models about how to model. The system thinking (Siegenfeld & Bar-Yam, 2020) is one of those. It adopts the viewpoint that the world consists of systems and subsystems. Each system is dynamic, depending on multiple interacting components. Harvey's spacetime dialectics (2008) reflect a similar perspective, emphasizing the space and time's relational and dynamic nature in shaping socio-economic processes. Thus, we can embed such a worldview into autonomous GIS: it first needs to identify the phenomena of interest as a function of a system and identify the positive and negative determinants. How will the higher-level system change the phenomena, and will the subsystems affect the higher-level system? Are the scales compatible? What are the time lags of the determinants? How do the determinants impose their forces through geographic space? What are the possible interactions between system components and system emergence? What is the leverage point for intervening the system? Gahegan (2020) has touched on some of these ideas, although not explicitly mentioned system thinking. Autonomous GIS should be able to consider these aspects when selecting models.

#### 5.3.3 Can autonomous GIS innovate?

In this paper, innovation refers to generating new research questions or solutions given particular contexts and constraints, such as creating new models or improving existing ones. Can AI generate novel ideas? GPT-4o has shown an advantage in low-price new product proposals (Meincke et al., 2024). Baek et al. (2024) explored the idea generation method starting from a seeding paper. Si et al. (2024) claimed that LLM-generated NLP research ideas are more novel than humans', although the feasibility and variety are weak. Hu et al. (2024) proposed an iterative approach to refine generated ideas to enhance novelty and diversity. Su et al. (2024) adopted a multi-agent system for idea generation. The agents pretend to be real scientists with different backgrounds to spark creative ideas through brainstorms. AI Scientist agent ( Lu et al., 2024) went further: it can generate novel research ideas, run experiments, and write full papers.

The abovementioned attempts demonstrate that LLMs can generate new ideas given the seed material (literature) and context (roles). We can provide seeds or incentives to LLM for innovation regarding autonomous GIS. For example, an autonomous GIS research agent can pretend to be a GIScience scholar based on real profiles and articles and then utilize its expertise to solve real-world problems reported in scientific or public news and stories. Considering the tremendous numbers of scholars and news reports, this might serve as an automatic way to generate massive research topics or solutions continuously, and many of them might connect fields that were rarely linked before due to limited personal backgrounds. These generated topics and solutions can then be discussed by a group of scholar agents and further reviewed by humans to determine which are worthy of further investigation, considering the limited resources. As autonomous GIS may bring large-scale automation, which could further promote innovations in GIScience in addition to technological advances. We advocate that GIS scientists use autonomy to facilitate innovation. For example, making geographic phenomena computable and then using rapid iterative variation and connection methods to discover patterns or solutions.

## 5.4 Impact on the GIScience community

Autonomous GIS may raise critical questions about trust, accountability, and its broader societal impact. Should we trust autonomous GIS? If it produces incorrect analyses or biased outputs, who is responsible? How does autonomous GIS influence traditional GIS workforce and education? This section provides brief discussions on these pressing questions.

### 5.4.1 Trustworthiness of autonomous GIS

Trust in autonomous GIS depends on its ability to explain its reasoning, provide verifiable outputs, and allow human oversight at critical decision points. Trustworthiness in this paper refers to the extent to which users can rely on the results from autonomous GIS. This concern is particularly relevant for individuals without a GIS background, as they must depend on the outputs of a complex "program" composed of one or more agents operating with various geospatial datasets that users may not fully understand. Conversely, GIS analysts can examine the transparent geoprocessing workflow, although autonomous GIS generates and executes it. They can also inspect the code or procedural steps by digging into the implementation details. However, transparency alone does not guarantee trustworthiness, as autonomous GIS may make mistakes similar to those of humans.

A community-driven approach would help develop an operationalizable evaluation framework to govern future GIS agents and ensure the validity of the generated results. For example, a comprehensive benchmarking framework or a dedicated reviewer agent could evaluate both the workflow and results to ensure their validity. Additionally, a human- or expert-in-the-loop problem-solving framework remains critical to ensure reliability in the autonomous GIS analysis and decision-making process. Ultimately, all users, whether GIS professionals or non-experts, remain responsible for interpreting and acting upon autonomous GIS outputs.

### 5.4.2 Embracing FAIR and AI-ready data for responsible autonomous GIS

The development of autonomous GIS and associated foundation models relies on various, high-quality geospatial datasets for training, benchmarking, and validation (Wang et al., 2024). Ensuring that these datasets adhere to the FAIR (Findable, Accessible, Interoperable, and Reusable) principles enhances their usability, longevity, and transparency, fostering responsible AI-powered spatial analysis (Tarboton et al., 2024). Initiatives such as the I-GUIDE Platform, supported by the U.S. National Science Foundation (NSF), have advanced FAIR data practices by providing cutting-edge AI, CyberGIS, infrastructure to support reproducible geospatial research, such as the Spatial AI Challenge (I-GUIDE, 2025; Michels et al., 2024; Spatial AI Challenge, 2024).

Considerations of social responsibility are particularly crucial in autonomous GIS applications (Janowicz, 2023), where decisions informed by AI models can impact critical areas such as urban planning, environmental monitoring, and disaster response (Jaroenchai et al., 2024). Ensuring that training data is representative and well-documented helps avoid hallucinations and mitigate potential harms while enhancing public trust in AI-powered GIS technologies. Therefore, AI-ready data are essential for autonomous GIS development. AI-ready data is not simply high-quality data; it must be representative of real-world variations, including patterns, outliers, and unexpected phenomena, to effectively train AI models. It must also ensure representativeness across regions and cultures. This requires structured formats, rich metadata, and interoperability to ensure that autonomous GIS can learn from and adapt to complex geospatial environments. However, building AI-ready geospatial datasets is challenging due to data heterogeneity, volume, labeling complexity, and privacy concerns. A key initiative addressing this is the NSF-led National Artificial Intelligence Research Resource (NAIRR) pilot, which aims to streamline access to AI-ready data, facilitating large-scale experimentation and ensuring responsible AI applications such as autonomous GIS. Model Concept Protocl, or MCP, is one AI-ready attempt to share the data and services with AI agents (Anthropic, 2024; X. Hou et al., 2025).

### 5.4.3 Human-AI collaboration and workforce adaptation in GIS

Given their increasing ability to perform relatively simple tasks, autonomous GIS agents may eventually reduce the need for entry-level GIS analysts. However, during the next few years, we expect such agents to act as augmentations that extend the abilities of GIS analysts rather than a replacement. In the long term, GIS agents may also take up tasks requiring more creativity and deeper domain understanding. To prepare for this shift, the GIS community should adapt professional training to incorporate collaborating with AI and autonomous agents into the curriculum and industry practice, while seeking methods to achieve the self-growing goal of autonomous GIS. Ideally, all future GIS analysts would understand fundamental AI principles, including the abilities and limitations of current AI systems.

Most importantly, geospatial analysts should be able to judge where human guidance and assessment are required. It is also likely that autonomous GIS agents will show a canonical behavior in approaching certain problems, favoring some over other methods, or recommending the same well-known datasets. While this is convenient both from a model training perspective and for performing routine downstream tasks, it is potentially problematic or even dangerous for rare tasks, in adapting to future needs and norms, and for creative problem-solving in general. Thus, educators will have to focus even more on demonstrating how to use AI constructively and critically and the ethical issues of the AI's consequential decision that may affect people's rights, for example, disenfranchising people's access to social services, such as health care and education (Consumer Protections for AI, 2024). Consideration of ethical issues is not only a good practice but is often a requirement for meeting governmental regulations such as the Colorado AI Act, C.R.S. § 6-1-1701 (Colorado General Assembly, 2024).

#### 5.4.4 Can the GIScience community take a leading role in shaping AI development?

Recent trends in AI research advocate the alignment of AI and the physical world (Sitti, 2021), expecting AI to comprehend the physics laws (Liu et al., 2025), 3D shapes of the surrounding scenes (Liu et al., 2024), and environmental sensor readings ( Xu et al., 2024). GIScience will potentially be essential in shaping AI advancements along this direction because the community has long been at the forefront of developing methods and technologies to enhance our understanding of the world. Specifically, geospatial research on the physical world focuses more on the tangible environment, such as physical geography and human-environment interaction, emphasizing the causation and association between geographic entities across various spatial scales and extents over time. Many impactful events to humans occur via the causal chain in the geospatial frame, such as population distribution and mobility, atmosphere dynamics, and land changes. GIScience has a tradition of investigating causation and association, and modeling is one important method for understanding these relationships. Thus, a question arises: can the GIScience community lead in shaping the development of AI that can understand the physical world, such as the causal chain of geographic phenomena?

This paper does not have a clear answer, although section 5.3 discusses autonomous modeling. One possible path may be developing a universal framework to model geographic phenomena using computational approaches. Generally, geographic entities, such as human settlements, jurisdictions, and natural resources, must transfer their influences to each other via geographic space. We may consider the friction of these influences as "distance relationship" and then project the geographic phenomenon into a high-dimensional space incorporating involved entities and their relationships. Next, we can apply computational methods similar to generative AI training to embed and predict geographic phenomena. If this LLM-like approach can generate seemingly reasonable results, we might say the model "understands" the physical world.

### 5.5 Critical research topics

In this section, we summarize the challenges and list the critical topics to chart the roadmap to autonomous GIS. Table 3 summarizes the research questions mentioned above and what we think will be important in the next five years from various aspects of autonomous goals, levels, functionality, scales, and impacts on the GIScience community.

**Table 3.** Critical research topics for autonomous GIS

| Aspect | Field | Research Topics |
|---|---|---|
| **Autonomous Goals** | Fine-tuning decision core | ○ Internal missing knowledge probe<br>○ Benchmark for autonomous GIS skills<br>○ Incorporating ethics, empathy, and holistic judgment into AI<br>○ Self-fine-tuning |
| | Self-growing decision core | ○ Enhancement of spatial programming skills via domain-specific feedback loops<br>○ Incremental and discrete self-growth via specialized agent fine-tuning<br>○ Co-growing models for human-AI symbiosis in GIS workflows |
| | Autonomous modeling | ○ Research paper replication by autonomous GIS<br>○ Environmental constrained autonomous modeling<br>○ Autonomous modeling for temporal dynamics and prediction<br>○ Modeling result interpretation<br>○ Meta-modeling |
| **Autonomous Levels** | Autonomous data preparation | ○ Autonomous online geospatial data discovery<br>○ Autonomous data collection<br>○ Geospatial data cleaning and integration |
| | Autonomous visualization | ○ Cartography agent<br>○ Autonomous interpretation from geo-visualization |
| | Memory-handling | ○ Identify and record valuable attempts<br>○ Logging mechanism for complex tasks |
| | Modeling refining | ○ Applicable geospatial models ranking for tasks<br>○ Result assessment and ranking |
| **Scales** | Multi-agent collaboration | ○ Collective learning<br>○ Inter-disciplinary agent collaboration<br>○ Autonomous GIS in real-time crisis management<br>○ Privacy considerations among users and agents |
| | Complex autonomous modeling | ○ Hierarchical task decomposing<br>○ Data and computational resources estimation for tasks |
| | Agent-led geographic research | ○ Research objective and hypothesis generation<br>○ Agentic AI for geospatial cyberinfrastructure |
| **Impacts on GIScience Community** | Privacy and Security | ○ User data privacy protection in model fine-tuning and collective learning<br>○ Personally identifiable information protection |
| | | ○ AI-ready geospatial data<br>○ Energy-aware GIS agents for sustainable spatial decision-making |
| | Trustworthiness and responsibility | ○ Uncertainty quantification, verification, and validation<br>○ Geo-ethical impact assessments<br>○ GIS professional education on agent-human collaboration<br>○ Human-centered and community-oriented perspectives |
| | Education | ○ Enhancing critical thinking in AI-assisted GIS tasks<br>○ Training GIS analysts in AI ethics.<br>○ Preparing students for AI-augmented GIS workflows<br>○ Promoting creativity in AI-powered GIS problem-solving |

## 6 Summary and Conclusion

The rise of generative AI provides a unique opportunity for GIScience, triggering the emergence and driving the evolution of autonomous GIS: a transformative paradigm where GIS can independently propose hypotheses, collect data, generate geoprocessing workflows, organize resources, execute spatial operations, verify results, and grow from learning operation experience for geographic research. We envision autonomous GIS as a next-generation AI-powered GIS, emerging as a new subfield within the broader domains of GIScience, spatial data sciene, and GeoAI (Li, 2020; Li, Arundel, et al., 2024). This vision, rooted in the five autonomous goals of self-generating, self-executing, self-verifying, self-organizing, and self-growing, represents the convergence of decades of GIS evolution and the emerging generative AI. Achieving these goals requires seamless integration of several core functions, including decision-making, data preparation, data operation, memory-handling, and core-updating, while maintaining ethical behavior.

The proposed autonomous levels, ranging from manual GIS (Level 0) to knowledge-aware GIS (Level 5), provide a structured pathway toward fully autonomous GIS. Current research has primarily advanced to workflow-aware GIS (Level 2), where systems can generate and execute workflows. Moving toward data-aware (Level 3), result-aware GIS (Level 4), and knowledge-aware GIS (Level 5) necessitates a focus on autonomous modeling to consider dynamic geospatial phenomena, adjust workflows based on results, and synthesize insights. Developing autonomous GIS across different scales (local, centralized, and infrastructure) also enables varying levels of computational and collaborative capabilities. Whereas local-scale systems focus on individual users and moderate tasks, centralized and infrastructure scales promise large-scale, interdisciplinary research advancements.

This paper outlines the concepts, architecture, research challenges, and agenda of autonomous GIS. Immediate objectives include developing benchmarks for assessing spatial modeling capabilities and enhancing AI's understanding of geospatial concepts. Long-term research might explore meta-modeling frameworks, dynamic and temporal modeling, and collaborative multi-agent systems. These efforts aim to create GIS approaches that not only process spatial data but also innovate and adapt through advanced modeling capabilities to understand complex geospatial dynamics better. Eventually, we hope that the human analysts and autonomous GIS can collaborate in creating efficient and accessible geospatial solutions at scale. We believe that by advancing the autonomy of GIS through the outlined goals, levels, functions, scales, and research agenda, geospatial researchers and industry will develop intelligent GIS systems capable of analyzing, predicting, reasoning, and adapting, and contribute to impactful scientific and societal advancements.

As geographers and GIScientists, we also believe it is essential to reflect not only on what is technically possible, but also on the societal and professional consequences of the current AI revolution, of which autonomous GIS forms a part. The automation of geospatial workflows could enhance GIS-based analysis and potentially help to remove human bias (intentional and unintentional) from the results produced. It could also help ensure that spatial data are used appropriately and make GIS technology easier to use across disciplines, sectors, and society at large. However, it also raises important questions: How can we be confident that autonomous systems do not themselves contain biases that we are unaware of? How will the growth in GIS automation affect the GIS profession? Will the demand for traditional GIS technical skills decline, reshaping the labor market for GIS professionals, or will it create new jobs centered on AI ethics and data governance? How will these changes influence GIS education, program enrollment, and the kinds of knowledge we prioritize in our current curricula? Conversely, what if the barriers to using GIS fall dramatically, and spatial analysis and modeling becomes ubiquitous and accessible to all? How might this catalyze new insights in research, improve governance, and empower civic engagement? We do not yet know the full implications, but we feel strongly that the GIScience community must actively engage with these questions to ensure we are shaping the future of (Geo) AI instead of being shaped by it. As we design and deploy increasingly intelligent geospatial systems, we carry a responsibility to ensure they are developed ethically, serve the public good, and support the continued value of human geographic insight in an AI-augmented future. What our research, and other similar efforts show, is that we must do this now. The mainstream use of AI in science is just around the corner, we have perhaps 2-5 years (Giattino et al., 2023; Boiko et al., 2023; Lu et al., 2024; Wang et al., 2023).

Finally, one aspect we have not discussed here, and that we leave for future work, is ensuring (super) alignment. Autonomous AI agents may not share the same representations of geography and society as we do; how can we ensure level 4 and 5 systems help to create a world that is in line with the *permanence of genuine human life on Earth* (Jonas, 1984)?

**Disclosure statement:** No potential conflict of interest was reported by the authors.

## Acknowledgments

We thank Professor May Yuan from the University of Texas at Dallas for her insightful comments and edits on an earlier version of the manuscript, and Professor Gengchen Mai from the University of Texas at Austin for sharing his insights at the 2025 AAG Autonomous GIS Panel. We also thank other team members of the Penn State Geoinformation and Big Data Research Lab for their contributions to this topic. Any use of trade, firm, or product names is for descriptive purposes only and does not imply endorsement by the U. S. Government.

## References

Agarwal, M., Sun, M., Kamath, C., Muslim, A., Sarker, P., Paul, J., Yee, H., Sieniek, M., Jablonski, K., Mayer, Y., Fork, D., Guia, S. de, McPike, J., Boulanger, A., Shekel, T., Schottlander, D., Xiao, Y., Manukonda, M. C., Liu, Y., … Prasad, G. (2024). *General Geospatial Inference with a Population Dynamics Foundation Model* (No. arXiv:2411.07207). arXiv. https://doi.org/10.48550/arXiv.2411.07207

Ageospatial. (2024, February 20). GeoForge: Geospatial Analysis with Large Language Models (GeoLLMs). *Medium*. https://medium.com/@ageospatial/geoforge-geospatial-analysis-with-large-language-models-geollms-2d3a0eaff8aa

Akinboyewa, T., Li, Z., Ning, H., & Lessani, M. N. (2024). *GIS Copilot: Towards an Autonomous GIS Agent for Spatial Analysis* (No. arXiv:2411.03205). arXiv. https://doi.org/10.48550/arXiv.2411.03205

Albrecht, C. M., Elmegreen, B., Gunawan, O., Hamann, H. F., Klein, L. J., Lu, S., Mariano, F., Siebenschuh, C., & Schmude, J. (2020). Next-generation geospatial-temporal information technologies for disaster management. *IBM Journal of Research and Development*, *64*(1/2), 5:1-5:12. IBM Journal of Research and Development. https://doi.org/10.1147/JRD.2020.2970903

Anselin, L. (1990). Spatial Dependence and Spatial Structural Instability in Applied Regression Analysis. *Journal of Regional Science*, *30*(2), 185–207. https://doi.org/10.1111/j.1467-9787.1990.tb00092.x

Anthropic. (2024, November 25). *Introducing the Model Context Protocol*. https://www.anthropic.com/news/model-context-protocol

Antti Karjalainen. (2024). *The Five Levels of Agentic Automation—Sema4.ai*. https://sema4.ai/blog/the-five-levels-of-agentic-automation/

Atil, B., Chittams, A., Fu, L., Ture, F., Xu, L., & Baldwin, B. (2024). *LLM Stability: A detailed analysis with some surprises* (No. arXiv:2408.04667). arXiv. https://doi.org/10.48550/arXiv.2408.04667

AutoGluon. (2025). *Autogluon/autogluon* [Python]. autogluon. https://github.com/autogluon/autogluon (Original work published 2019)

Baek, J., Jauhar, S. K., Cucerzan, S., & Hwang, S. J. (2024). *ResearchAgent: Iterative Research Idea Generation over Scientific Literature with Large Language Models* (No. arXiv:2404.07738). arXiv. https://doi.org/10.48550/arXiv.2404.07738

Bi, J., Zhang, X., Yuan, H., Zhang, J., & Zhou, M. (2022). A Hybrid Prediction Method for Realistic Network Traffic With Temporal Convolutional Network and LSTM. *IEEE Transactions on Automation Science and Engineering*, *19*(3), 1869–1879. IEEE Transactions on Automation Science and Engineering. https://doi.org/10.1109/TASE.2021.3077537

Boiko, D. A., MacKnight, R., Kline, B., & Gomes, G. (2023). Autonomous chemical research with large language models. *Nature*, *624*(7992), 570–578.

Bryson, T. (2024, November 14). *From questions to discoveries: NASA's new Earth Copilot brings Microsoft AI capabilities to democratize access to complex data*. The Official Microsoft Blog. https://blogs.microsoft.com/blog/2024/11/14/from-questions-to-discoveries-nasas-new-earth-copilot-brings-microsoft-ai-capabilities-to-democratize-access-to-complex-data/

Cail Smith. (2016). *Turning Transportation: Challenges and Opportunities Presented to the City of Vancouver by Autonomous Vehicles*. https://sustain.ubc.ca/about/resources/turning-transportation-challenges-and-opportunities-presented-city-vancouver

Chen, Y., Wang, W., Lobry, S., & Kurtz, C. (2024). *An LLM Agent for Automatic Geospatial Data Analysis* (No. arXiv:2410.18792). arXiv. https://doi.org/10.48550/arXiv.2410.18792

Chen, Y.-J., Li, W., Losier, L.-M., Tucker, C., Janowicz, K., Chou, T.-Y. (Jimmy), Shinde, R., Reichardt, M., Gupta, A. K., Durbha, S., Chauhan, L., Maskey, M., Simonis, I., Sen, S., Churchyard, P. A., & Simmons, S. (2025). *Thirty Years of the Open Geospatial Consortium—History, Present, and Future*. https://eartharxiv.org/repository/view/8623/

Chui, M., Hazan, E., Roberts, R., Singla, A., & Smaje, K. (2023). *The economic potential of generative AI*. http://dln.jaipuria.ac.in:8080/jspui/bitstream/123456789/14313/1/The-economic-potential-of-generative-ai-the-next-productivity-frontier.pdf

Colorado AI Act - Consumer Protections for Artificial Intelligence, SB24-205, Colorado General Assembly 2024 Regular Session (2024). https://leg.colorado.gov/sites/default/files/images/colorado_ai_act_-_public_interest_groups_redline_2024.10.30.pdf

Cook, J. (2024). *OpenAI's 5 Levels Of 'Super AI' (AGI To Outperform Human Capability)*. Forbes. https://www.forbes.com/sites/jodiecook/2024/07/16/openais-5-levels-of-super-ai-agi-to-outperform-human-capability/

David Schottlander & Tomer Shekel. (2025, April 8). *Geospatial Reasoning: Unlocking insights with generative AI and multiple foundation models*. https://research.google/blog/geospatial-reasoning-unlocking-insights-with-generative-ai-and-multiple-foundation-models/

DeepSeek-AI, Liu, A., Feng, B., Xue, B., Wang, B., Wu, B., Lu, C., Zhao, C., Deng, C., Zhang, C., Ruan, C., Dai, D., Guo, D., Yang, D., Chen, D., Ji, D., Li, E., Lin, F., Dai, F., … Pan, Z. (2024). *DeepSeek-V3 Technical Report* (No. arXiv:2412.19437; Version 1). arXiv. https://doi.org/10.48550/arXiv.2412.19437

Epstein, Z., Hertzmann, A., & the Investigators of Human Creativity. (2023). Art and the science of generative AI. *Science*, *380*(6650), 1110–1111. https://doi.org/10.1126/science.adh4451

Esri. (2025). *History of GIS | Timeline of the Development of GIS*. https://www.esri.com/en-us/what-is-gis/history-of-gis

Fan, C., Yang, Y., & Mostafavi, A. (2024). Neural embeddings of urban big data reveal spatial structures in cities. *Humanities and Social Sciences Communications*, *11*(1), 1–15. https://doi.org/10.1057/s41599-024-02917-6

Fanfarillo, A., Roozitalab, B., Hu, W., & Cervone, G. (2021). Probabilistic forecasting using deep generative models. *GeoInformatica*, *25*(1), 127–147. https://doi.org/10.1007/s10707-020-00425-8

Ferrara, E. (2023). Should ChatGPT be Biased? Challenges and Risks of Bias in Large Language Models. *First Monday*. https://doi.org/10.5210/fm.v28i11.13346

Feuerriegel, S., Hartmann, J., Janiesch, C., & Zschech, P. (2024). Generative AI. *Business & Information Systems Engineering*, *66*(1), 111–126. https://doi.org/10.1007/s12599-023-00834-7

Feurer, M., Eggensperger, K., Falkner, S., Lindauer, M., & Hutter, F. (2022). Auto-Sklearn 2.0: Hands-free AutoML via Meta-Learning. *Journal of Machine Learning Research*, *23*(261), 1–61.

Fortnow, L. (2009). The status of the P versus NP problem. *Communications of the ACM*, *52*(9), 78–86. https://doi.org/10.1145/1562164.1562186

Gahegan, M. (2020). Fourth paradigm GIScience? Prospects for automated discovery and explanation from data. *International Journal of Geographical Information Science*, *34*(1), 1–21. https://doi.org/10.1080/13658816.2019.1652304

Gahegan, M., & Adams, B. (2014). Re-Envisioning Data Description Using Peirce's Pragmatics. In M. Duckham, E. Pebesma, K. Stewart, & A. U. Frank (Eds.), *Geographic Information Science* (pp. 142–158). Springer International Publishing. https://doi.org/10.1007/978-3-319-11593-1_10

Gallegos, I. O., Rossi, R. A., Barrow, J., Tanjim, M. M., Kim, S., Dernoncourt, F., Yu, T., Zhang, R., & Ahmed, N. K. (2024). Bias and Fairness in Large Language Models: A Survey. *Computational Linguistics*, *50*(3), 1097–1179. https://doi.org/10.1162/coli_a_00524

Garikapati, D., & Shetiya, S. S. (2024). Autonomous Vehicles: Evolution of Artificial Intelligence and the Current Industry Landscape. *Big Data and Cognitive Computing*, *8*(4), Article 4. https://doi.org/10.3390/bdcc8040042

Giattino, C., Mathieu, E., Samborska, V., & Roser, M. (2023). Artificial Intelligence. *Our World in Data*. https://ourworldindata.org/artificial-intelligence

Goodchild, M. F. (1992). Geographical information science. *International Journal of Geographical Information Systems*, *6*(1), 31–45. https://doi.org/10.1080/02693799208901893

Goodchild, M. F., & Li, W. (2021). Replication across space and time must be weak in the social and environmental sciences. *Proceedings of the National Academy of Sciences*, *118*(35), e2015759118. https://doi.org/10.1073/pnas.2015759118

Google LLC. (2024). *Data Science Agent*. https://labs.google.com/code/dsa

Goonewardena, K. (2023). David Harvey, geography and Marxism. *Scottish Geographical Journal*, *139*(3–4), 415–436. https://doi.org/10.1080/14702541.2023.2260824

Gorelick, N., Hancher, M., Dixon, M., Ilyushchenko, S., Thau, D., & Moore, R. (2017). Google Earth Engine: Planetary-scale geospatial analysis for everyone. *Remote Sensing of Environment*, *202*, 18–27.

Greyling, C. (2024, October 11). 5 Levels Of AI Agents (Updated). *Medium*. https://cobusgreyling.medium.com/5-levels-of-ai-agents-updated-0ddf8931a1c6

Guo, H., Su, X., Wu, C., Du, B., Zhang, L., & Li, D. (2024). *Remote Sensing ChatGPT: Solving Remote Sensing Tasks with ChatGPT and Visual Models* (No. arXiv:2401.09083). arXiv. https://doi.org/10.48550/arXiv.2401.09083

Guo, S., Deng, C., Wen, Y., Chen, H., Chang, Y., & Wang, J. (2024). *DS-Agent: Automated Data Science by Empowering Large Language Models with Case-Based Reasoning* (No. arXiv:2402.17453). arXiv. https://doi.org/10.48550/arXiv.2402.17453

Gupta, D. V., Ishaqui, A. S. A., & Kadiyala, D. K. (2024). *Geode: A Zero-shot Geospatial Question-Answering Agent with Explicit Reasoning and Precise Spatio-Temporal Retrieval* (No. arXiv:2407.11014). arXiv. https://doi.org/10.48550/arXiv.2407.11014

Gupta, J., Molnar, C., Xien, Y., Knight, J., & Shekhar, S. (2021). Spatial Variability Aware Deep Neural Networks (SVANN): A General Approach. *ACM Transactions on Intelligent Systems and Technology (TIST)*. https://doi.org/10.1145/3466688

Gupta, P., & Gahegan, M. (2020). Categories are in flux, but their computational representations are fixed: That's a problem. *Transactions in GIS*, *24*(2), 291–314. https://doi.org/10.1111/tgis.12602

Hanover, D., Loquercio, A., Bauersfeld, L., Romero, A., Penicka, R., Song, Y., Cioffi, G., Kaufmann, E., & Scaramuzza, D. (2024). Autonomous Drone Racing: A Survey. *IEEE Transactions on Robotics*, *40*, 3044–3067. IEEE Transactions on Robotics. https://doi.org/10.1109/TRO.2024.3400838

Harvey, D. (2008). The Dialectics of Spacetime. In B. Ollman & T. Smith (Eds.), *Dialectics for the New Century* (pp. 98–117). Palgrave Macmillan UK. https://doi.org/10.1057/9780230583818_7

Hong, D., Zhang, B., Li, X., Li, Y., Li, C., Yao, J., Yokoya, N., Li, H., Ghamisi, P., & Jia, X. (2024). SpectralGPT: Spectral remote sensing foundation model. *IEEE Transactions on Pattern Analysis and Machine Intelligence*. https://ieeexplore.ieee.org/abstract/document/10490262/?casa_token=44HT_Icef2gAAAAA:Qr0IGIT39mLFV6TgYdBIfE8Igkd9USPrRnw_M91VPLP34z6FWlbKNy1q4kEXLkOwCc5-GPw

Hong, S., Lin, Y., Liu, B., Liu, B., Wu, B., Zhang, C., Wei, C., Li, D., Chen, J., Zhang, J., Wang, J., Zhang, L., Zhang, L., Yang, M., Zhuge, M., Guo, T., Zhou, T., Tao, W., Tang, X., … Wu, C. (2024). *Data Interpreter: An LLM Agent For Data Science* (No. arXiv:2402.18679). arXiv. https://doi.org/10.48550/arXiv.2402.18679

Hou, S., Shen, Z., Zhao, A., Liang, J., Gui, Z., Guan, X., Li, R., & Wu, H. (2024). *GeoCode-GPT: A Large Language Model for Geospatial Code Generation Tasks* (No. arXiv:2410.17031). arXiv. https://doi.org/10.48550/arXiv.2410.17031

Hou, X., Zhao, Y., Wang, S., & Wang, H. (2025). *Model Context Protocol (MCP): Landscape, Security Threats, and Future Research Directions* (No. arXiv:2503.23278). arXiv. https://doi.org/10.48550/arXiv.2503.23278

Hsu, C.-Y., Li, W., & Wang, S. (2024). Geospatial foundation models for image analysis: Evaluating and enhancing NASA-IBM Prithvi's domain adaptability. *International Journal of Geographical Information Science*, 1–30. https://doi.org/10.1080/13658816.2024.2397441

Hu, E. J., Shen, Y., Wallis, P., Allen-Zhu, Z., Li, Y., Wang, S., Wang, L., & Chen, W. (2021). *LoRA: Low-Rank Adaptation of Large Language Models* (No. arXiv:2106.09685). arXiv. https://doi.org/10.48550/arXiv.2106.09685

Hu, X., Fu, H., Wang, J., Wang, Y., Li, Z., Xu, R., Lu, Y., Jin, Y., Pan, L., & Lan, Z. (2024). *Nova: An Iterative Planning and Search Approach to Enhance Novelty and Diversity of LLM Generated Ideas* (No. arXiv:2410.14255). arXiv. https://doi.org/10.48550/arXiv.2410.14255

Huang, C., Chen, S., Li, Z., Qu, J., Xiao, Y., Liu, J., & Chen, Z. (2024). GeoAgent: To Empower LLMs using Geospatial Tools for Address Standardization. In L.-W. Ku, A. Martins, & V. Srikumar (Eds.), *Findings of the Association for Computational Linguistics: ACL 2024* (pp. 6048–6063). Association for Computational Linguistics. https://doi.org/10.18653/v1/2024.findings-acl.362

Huang, H. (2022). Location Based Services. In W. Kresse & D. Danko (Eds.), *Springer Handbook of Geographic Information* (pp. 629–637). Springer International Publishing. https://doi.org/10.1007/978-3-030-53125-6_22

Huang, Y. (2024). *Levels of AI Agents: From Rules to Large Language Models* (No. arXiv:2405.06643). arXiv. https://doi.org/10.48550/arXiv.2405.06643

Huang, Y., Gao, T., Xu, H., Zhao, Q., Song, Y., Gui, Z., Lv, T., Chen, H., Cui, L., Li, S., & Wei, F. (2025). *PEACE: Empowering Geologic Map Holistic Understanding with MLLMs* (No. arXiv:2501.06184). arXiv. https://doi.org/10.48550/arXiv.2501.06184

I-GUIDE. (2025). *I-GUIDE Platform*. https://platform.i-guide.io/

Jakubik, J., Roy, S., Phillips, C. E., Fraccaro, P., Godwin, D., Zadrozny, B., Szwarcman, D., Gomes, C., Nyirjesy, G., Edwards, B., Kimura, D., Simumba, N., Chu, L., Mukkavilli, S. K., Lambhate, D., Das, K., Bangalore, R., Oliveira, D., Muszynski, M., … Ramachandran, R. (2023). *Foundation Models for Generalist Geospatial Artificial Intelligence* (No. arXiv:2310.18660). arXiv. https://doi.org/10.48550/arXiv.2310.18660

Janowicz, K. (2023). Philosophical Foundations of GeoAI: Exploring Sustainability, Diversity, and Bias in GeoAI and Spatial Data Science. In *Handbook of Geospatial Artificial Intelligence*. CRC Press.

Janowicz, K., Gao, S., McKenzie, G., Hu, Y., & Bhaduri, B. (2020). GeoAI: Spatially explicit artificial intelligence techniques for geographic knowledge discovery and beyond. *International Journal of Geographical Information Science*, *34*(4), 625–636. https://doi.org/10.1080/13658816.2019.1684500

Jaroenchai, N., Wang, S., Stanislawski, L. V., Shavers, E., Jiang, Z., Sagan, V., & Usery, E. L. (2024). Transfer learning with convolutional neural networks for hydrological streamline delineation. *Environmental Modelling & Software*, *181*, 106165. https://doi.org/10.1016/j.envsoft.2024.106165

Jonas, H. (1984). *The Imperative of Responsibility: In Search of an Ethics for the Technological Age*. University of Chicago Press. https://press.uchicago.edu/ucp/books/book/chicago/I/bo5953283.html

Kaddar, A., Abta, A., & Alaoui, H. T. T. (2011). A comparison of delayed SIR and SEIR epidemic models. *Nonlinear Analysis: Modelling and Control*, *16*(2), Article 2. https://doi.org/10.15388/NA.16.2.14104

Kaddour, J., Harris, J., Mozes, M., Bradley, H., Raileanu, R., & McHardy, R. (2023). *Challenges and Applications of Large Language Models* (No. arXiv:2307.10169). arXiv. https://doi.org/10.48550/arXiv.2307.10169

Kaplan, J., McCandlish, S., Henighan, T., Brown, T. B., Chess, B., Child, R., Gray, S., Radford, A., Wu, J., & Amodei, D. (2020). *Scaling Laws for Neural Language Models* (No. arXiv:2001.08361). arXiv. https://doi.org/10.48550/arXiv.2001.08361

K.c., U., Garg, S., Hilton, J., Aryal, J., & Forbes-Smith, N. (2019). Cloud Computing in natural hazard modeling systems: Current research trends and future directions. *International Journal of Disaster Risk Reduction*, *38*, 101188. https://doi.org/10.1016/j.ijdrr.2019.101188

Kearns, F. R., Kelly, M., & Tuxen, K. A. (2003). Everything happens somewhere: Using webGIS as a tool for sustainable natural resource management. *Frontiers in Ecology and the Environment*, *1*(10), 541–548. https://doi.org/10.1890/1540-9295(2003)001[0541:EHSUWA]2.0.CO;2

Kholoshyn, I., Bondarenko, O., Hanchuk, O., & Shmeltser, E. (2019). *Cloud ArcGIS Online as an innovative tool for developing geoinformation competence with future geography teachers* (No. arXiv:1909.04388). arXiv. https://doi.org/10.48550/arXiv.1909.04388

Khune, A. (2024, May 30). *Modernizing Uber's Batch Data Infrastructure with Google Cloud Platform*. Uber Blog. https://www.uber.com/blog/modernizing-ubers-data-infrastructure-with-gcp/

Kim, J., Mishra, A. K., Limosani, R., Scafuro, M., Cauli, N., Santos-Victor, J., Mazzolai, B., & Cavallo, F. (2019). Control strategies for cleaning robots in domestic applications: A comprehensive review. *International Journal of Advanced Robotic Systems*, *16*(4), 1729881419857432. https://doi.org/10.1177/1729881419857432

Kommineni, V. K., König-Ries, B., & Samuel, S. (2024). *From human experts to machines: An LLM supported approach to ontology and knowledge graph construction* (No. arXiv:2403.08345). arXiv. https://doi.org/10.48550/arXiv.2403.08345

Kononykhin, D., Mozikov, M., Mishtal, K., Kuznetsov, P., Abramov, D., Sotiriadi, N., Maximov, Y., Savchenko, A. V., & Makarov, I. (2024). From Data to Decisions: Streamlining Geospatial Operations with Multimodal GlobeFlowGPT. *Proceedings of the 32nd ACM International Conference on Advances in Geographic Information Systems*, 649–652. https://doi.org/10.1145/3678717.3691248

Kwan, M.-P. (2012). The Uncertain Geographic Context Problem. *Annals of the Association of American Geographers*, *102*(5), 958–968. https://doi.org/10.1080/00045608.2012.687349

L. C. M., Barrault, L., Duquenne, P.-A., Elbayad, M., Kozhevnikov, A., Alastruey, B., Andrews, P., Coria, M., Couairon, G., Costa-jussà, M. R., Dale, D., Elsahar, H., Heffernan, K., Janeiro, J. M., Tran, T., Ropers, C., Sánchez, E., Roman, R. S., Mourachko, A., … Schwenk, H. (2024). *Large Concept Models: Language Modeling in a Sentence Representation Space* (No. arXiv:2412.08821). arXiv. https://doi.org/10.48550/arXiv.2412.08821

Lewis, P., Perez, E., Piktus, A., Petroni, F., Karpukhin, V., Goyal, N., Küttler, H., Lewis, M., Yih, W., Rocktäschel, T., Riedel, S., & Kiela, D. (2020). Retrieval-Augmented Generation for Knowledge-Intensive NLP Tasks. *Advances in Neural Information Processing Systems*, *33*, 9459–9474. https://proceedings.neurips.cc/paper/2020/hash/6b493230205f780e1bc26945df7481e5-Abstract.html

Li, W. (2018). Lowering the Barriers for Accessing Distributed Geospatial Big Data to Advance Spatial Data Science: The PolarHub Solution. *Annals of the American Association of Geographers*, *108*(3), 773–793. https://doi.org/10.1080/24694452.2017.1373625

Li, W. (2020). GeoAI: Where machine learning and big data converge in GIScience. *Journal of Spatial Information Science*, *20*, Article 20.

Li, W., Arundel, S., Gao, S., Goodchild, M., Hu, Y., Wang, S., & Zipf, A. (2024). GeoAI for Science and the Science of GeoAI. *Journal of Spatial Information Science*, *29*, Article 29. https://doi.org/10.5311/JOSIS.2024.29.349

Li, W., Hsu, C.-Y., Wang, S., Yang, Y., Lee, H., Liljedahl, A., Witharana, C., Yang, Y., Rogers, B. M., Arundel, S. T., Jones, M. B., McHenry, K., & Solis, P. (2024). Segment Anything Model Can Not Segment Anything: Assessing AI Foundation Model's Generalizability in Permafrost Mapping. *Remote Sensing*, *16*(5), Article 5. https://doi.org/10.3390/rs16050797

Li, W., Li, L., Goodchild, M. F., & Anselin, L. (2013). A Geospatial Cyberinfrastructure for Urban Economic Analysis and Spatial Decision-Making. *ISPRS International Journal of Geo-Information*, *2*(2), Article 2. https://doi.org/10.3390/ijgi2020413

Li, Z., & Ning, H. (2023). Autonomous GIS: The next-generation AI-powered GIS. *International Journal of Digital Earth*, *16*(2), 4668–4686. https://doi.org/10.1080/17538947.2023.2278895

Li, Z., Qiao, S., Jiang, Y., & Li, X. (2021). Building a social media-based HIV risk behavior index to inform the prediction of HIV new diagnosis: A feasibility study. *AIDS (London, England)*, *35*(Suppl 1), S91–S99. https://doi.org/10.1097/QAD.0000000000002787

Li, Z., Zang, Q., Ma, D., Guo, J., Zheng, T., Liu, M., Niu, X., Wang, Y., Yang, J., Liu, J., Zhong, W., Zhou, W., Huang, W., & Zhang, G. (2024). *AutoKaggle: A Multi-Agent Framework for Autonomous Data Science Competitions* (No. arXiv:2410.20424). arXiv. https://doi.org/10.48550/arXiv.2410.20424

Lime, S. (2008). MapServer. In G. B. Hall & M. G. Leahy (Eds.), *Open Source Approaches in Spatial Data Handling* (Vol. 2, pp. 65–85). Springer Berlin Heidelberg. https://doi.org/10.1007/978-3-540-74831-1_4

Lin, L., Wang, L., Guo, J., & Wong, K.-F. (2024). *Investigating Bias in LLM-Based Bias Detection: Disparities between LLMs and Human Perception* (No. arXiv:2403.14896). arXiv. https://doi.org/10.48550/arXiv.2403.14896

Lin, Q., Hu, R., Li, H., Wu, S., Li, Y., Fang, K., Feng, H., Du, Z., & Xu, L. (2024). *ShapefileGPT: A Multi-Agent Large Language Model Framework for Automated Shapefile Processing* (No. arXiv:2410.12376). arXiv. https://doi.org/10.48550/arXiv.2410.12376

Liu, C., Chen, K., Zhang, H., Qi, Z., Zou, Z., & Shi, Z. (2024). Change-Agent: Towards Interactive Comprehensive Remote Sensing Change Interpretation and Analysis. *IEEE Transactions on Geoscience and Remote Sensing*, *62*, 1–16. https://doi.org/10.1109/TGRS.2024.3425815

Liu, D., Zhang, J., Dinh, A.-D., Park, E., Zhang, S., & Xu, C. (2025). *Generative Physical AI in Vision: A Survey* (No. arXiv:2501.10928). arXiv. https://doi.org/10.48550/arXiv.2501.10928

Liu, Y., Chen, W., Bai, Y., Liang, X., Li, G., Gao, W., & Lin, L. (2024). *Aligning Cyber Space with Physical World: A Comprehensive Survey on Embodied AI* (No. arXiv:2407.06886). arXiv. https://doi.org/10.48550/arXiv.2407.06886

Lu, C., Lu, C., Lange, R. T., Foerster, J., Clune, J., & Ha, D. (2024). *The AI Scientist: Towards Fully Automated Open-Ended Scientific Discovery* (No. arXiv:2408.06292). arXiv. https://doi.org/10.48550/arXiv.2408.06292

Lu, J., Liu, A., Dong, F., Gu, F., Gama, J., & Zhang, G. (2018). Learning under concept drift: A review. *IEEE Transactions on Knowledge and Data Engineering*, *31*(12), 2346–2363.

Lu, S., Guo, J., Zimmer-Dauphinee, J. R., Nieusma, J. M., Wang, X., VanValkenburgh, P., Wernke, S. A., & Huo, Y. (2024). *AI Foundation Models in Remote Sensing: A Survey* (No. arXiv:2408.03464). arXiv. https://doi.org/10.48550/arXiv.2408.03464

Lyu, F., Wang, S., Han, S. Y., Catlett, C., & Wang, S. (2022). An integrated cyberGIS and machine learning framework for fine-scale prediction of Urban Heat Island using satellite remote sensing and urban sensor network data. *Urban Informatics*, *1*(1), 6. https://doi.org/10.1007/s44212-022-00002-4

Mahdi Farnaghi, Gustavo García, & Zehao Lu. (2024, December). *IntelliGeo*. https://www.intelligeo.org/

Mai, G., Huang, W., Sun, J., Song, S., Mishra, D., Liu, N., Gao, S., Liu, T., Cong, G., Hu, Y., Cundy, C., Li, Z., Zhu, R., & Lao, N. (2024). On the Opportunities and Challenges of Foundation Models for GeoAI (Vision Paper). *ACM Trans. Spatial Algorithms Syst.*, *10*(2), 11:1-11:46. https://doi.org/10.1145/3653070

Majic, I., Wang, Z., Janowicz, K., & Karimi, M. (2024). Spatial Task-Explicity Matters in Prompting Large Multimodal Models for Spatial Planning. *Proceedings of the 7th ACM SIGSPATIAL International Workshop on AI for Geographic Knowledge Discovery*, 99–105. https://doi.org/10.1145/3687123.3698293

Makke, N., & Chawla, S. (2024). Interpretable scientific discovery with symbolic regression: A review. *Artificial Intelligence Review*, *57*(1), 2. https://doi.org/10.1007/s10462-023-10622-0

Mansourian, A., & Oucheikh, R. (2024). ChatGeoAI: Enabling Geospatial Analysis for Public through Natural Language, with Large Language Models. *ISPRS International Journal of Geo-Information*, *13*(10), Article 10. https://doi.org/10.3390/ijgi13100348

Manus AI. (2025). *Manus*. https://manus.im/

MASSER, I. (1999). All shapes and sizes: The first generation of national spatial data infrastructures. *International Journal of Geographical Information Science*, *13*(1), 67–84. https://doi.org/10.1080/136588199241463

Meincke, L., Girotra, K., Nave, G., Terwiesch, C., & Ulrich, K. T. (2024). *Using Large Language Models for Idea Generation in Innovation* (SSRN Scholarly Paper No. 4526071). Social Science Research Network. https://doi.org/10.2139/ssrn.4526071

Memduhoğlu, A. (2025). Towards AI-Assisted Mapmaking: Assessing the Capabilities of GPT-4o in Cartographic Design. *ISPRS International Journal of Geo-Information*, *14*(1), Article 1. https://doi.org/10.3390/ijgi14010035

Michels, A. C., Padmanabhan, A., Xiao, Z., Kotak, M., Baig, F., & Wang, S. (2024). CyberGIS-Compute: Middleware for democratizing scalable geocomputation. *SoftwareX*, *26*, 101691. https://doi.org/10.1016/j.softx.2024.101691

Michels, A., Padmanabhan, A., Li, Z., & Wang, S. (2024). EasyScienceGateway: A new framework for providing reproducible user environments on science gateways. *Concurrency and Computation: Practice and Experience*, *36*(4), e7929. https://doi.org/10.1002/cpe.7929

Mooney, P., & Minghini, M. (2017). A review of OpenStreetMap data. *Mapping and the Citizen Sensor*, 37–59.

Morris, M. R., Sohl-dickstein, J., Fiedel, N., Warkentin, T., Dafoe, A., Faust, A., Farabet, C., & Legg, S. (2024). *Levels of AGI for Operationalizing Progress on the Path to AGI* (No. arXiv:2311.02462). arXiv. https://doi.org/10.48550/arXiv.2311.02462

Nestler, J. (2024, February 14). How to find geometry intersections within the same dataset using geopandas. *Medium*. https://medium.com/@jesse.b.nestler/how-to-find-geometry-intersections-within-the-same-dataset-using-geopandas-59cd1a5f30f9

Ning, H., Li, Z., Akinboyewa, T., & Lessani, M. N. (2024). *An Autonomous GIS Agent Framework for Geospatial Data Retrieval* (No. arXiv:2407.21024). arXiv. http://arxiv.org/abs/2407.21024

Ning, H., Li, Z., Akinboyewa, T., & Lessani, M. N. (2025). An autonomous GIS agent framework for geospatial data retrieval. *International Journal of Digital Earth*, *18*(1), 2458688. https://doi.org/10.1080/17538947.2025.2458688

Norman, S. P., Hargrove, W. W., Spruce, J. P., Christie, W. M., & Schroeder, S. W. (2013). *Using the ForWarn System*. https://research.fs.usda.gov/sites/default/files/2023-02/srs-efetac_forwarn_gtr180.pdf

Olson, R. S., & Moore, J. H. (2016). TPOT: A Tree-based Pipeline Optimization Tool for Automating Machine Learning. *Proceedings of the Workshop on Automatic Machine Learning*, 66–74. https://proceedings.mlr.press/v64/olson_tpot_2016.html

Österman, A. (2014). *Map visualization in ArcGIS, QGIS and MapInfo*. https://www.diva-portal.org/smash/record.jsf?pid=diva2:729183

Pagnoni, A., Pasunuru, R., Rodriguez, P., Nguyen, J., Muller, B., Li, M., Zhou, C., Yu, L., Weston, J., Zettlemoyer, L., Ghosh, G., Lewis, M., Holtzman, A., & Iyer, S. (2024). *Byte Latent Transformer: Patches Scale Better Than Tokens* (No. arXiv:2412.09871). arXiv. https://doi.org/10.48550/arXiv.2412.09871

Peng, L., Giampouras, P., & Vidal, R. (2023). The Ideal Continual Learner: An Agent That Never Forgets. *Proceedings of the 40th International Conference on Machine Learning*, 27585–27610. https://proceedings.mlr.press/v202/peng23a.html

Rao, J., Gao, S., Mai, G., & Janowicz, K. (2023). Building Privacy-Preserving and Secure Geospatial Artificial Intelligence Foundation Models (Vision Paper). *Proceedings of the 31st ACM International Conference on Advances in Geographic Information Systems*, 1–4. https://doi.org/10.1145/3589132.3625611

Salehin, I., Islam, Md. S., Saha, P., Noman, S. M., Tuni, A., Hasan, Md. M., & Baten, Md. A. (2024). AutoML: A systematic review on automated machine learning with neural architecture search. *Journal of Information and Intelligence*, *2*(1), 52–81. https://doi.org/10.1016/j.jiixd.2023.10.002

Schaeffer, R., Miranda, B., & Koyejo, S. (2023). *Are Emergent Abilities of Large Language Models a Mirage?* (No. arXiv:2304.15004). arXiv. https://doi.org/10.48550/arXiv.2304.15004

Schmidgall, S., Su, Y., Wang, Z., Sun, X., Wu, J., Yu, X., Liu, J., Liu, Z., & Barsoum, E. (2025). *Agent Laboratory: Using LLM Agents as Research Assistants* (No. arXiv:2501.04227). arXiv. https://doi.org/10.48550/arXiv.2501.04227

Sengar, S. S., Hasan, A. B., Kumar, S., & Carroll, F. (2024). Generative artificial intelligence: A systematic review and applications. *Multimedia Tools and Applications*. https://doi.org/10.1007/s11042-024-20016-1

Shalizi, C. R. (2006). Methods and Techniques of Complex Systems Science: An Overview. In T. S. Deisboeck & J. Y. Kresh (Eds.), *Complex Systems Science in Biomedicine* (pp. 33–114). Springer US. https://doi.org/10.1007/978-0-387-33532-2_2

Shi, M., Janowicz, K., Verstegen, J., Currier, K., Wiedemann, N., Mai, G., Majic, I., Liu, Z., & Zhu, R. (2025). Geography for AI Sustainability and Sustainability for GeoAI. *Cartography and Geographic Information Science (in-Press)*.

Si, C., Yang, D., & Hashimoto, T. (2024). *Can LLMs Generate Novel Research Ideas? A Large-Scale Human Study with 100+ NLP Researchers* (No. arXiv:2409.04109). arXiv. https://doi.org/10.48550/arXiv.2409.04109

Siegenfeld, A. F., & Bar-Yam, Y. (2020). An Introduction to Complex Systems Science and Its Applications. *Complexity*, *2020*, e6105872. https://doi.org/10.1155/2020/6105872

Singh, S., Fore, M., & Stamoulis, D. (2024). *GeoLLM-Engine: A Realistic Environment for Building Geospatial Copilots* (No. arXiv:2404.15500). arXiv. https://doi.org/10.48550/arXiv.2404.15500

Sitti, M. (2021). Physical intelligence as a new paradigm. *Extreme Mechanics Letters*, *46*, 101340. https://doi.org/10.1016/j.eml.2021.101340

Spatial AI Challenge. (2024). *Spatial AI Challenge 2024: Fostering FAIR Data and Open Science Practices Using the I-GUIDE Platform: I-GUIDE*. https://i-guide.io/spatial-ai-challenge-2024/

Starace, L. L. L., & Di Martino, S. (2024). Can Large Language Models Automatically Generate GIS Reports? In M. Lotfian & L. L. L. Starace (Eds.), *Web and Wireless Geographical Information Systems* (pp. 147–161). Springer Nature Switzerland. https://doi.org/10.1007/978-3-031-60796-7_11

Su, H., Chen, R., Tang, S., Zheng, X., Li, J., Yin, Z., Ouyang, W., & Dong, N. (2024). *Two Heads Are Better Than One: A Multi-Agent System Has the Potential to Improve Scientific Idea Generation* (No. arXiv:2410.09403). arXiv. https://doi.org/10.48550/arXiv.2410.09403

Synopsys. (2019, April 14). *The 6 Levels of Vehicle Autonomy Explained | Synopsys Automotive*. https://www.synopsys.com/blogs/chip-design/autonomous-driving-levels.html

Szwarcman, D., Roy, S., Fraccaro, P., Gíslason, Þ. E., Blumenstiel, B., Ghosal, R., Oliveira, P. H. de, Almeida, J. L. de S., Sedona, R., Kang, Y., Chakraborty, S., Wang, S., Kumar, A., Truong, M., Godwin, D., Lee, H., Hsu, C.-Y., Asanjan, A. A., Mujeci, B., … Moreno, J. B. (2024). *Prithvi-EO-2.0: A Versatile Multi-Temporal Foundation Model for Earth Observation Applications* (No. arXiv:2412.02732). arXiv. https://doi.org/10.48550/arXiv.2412.02732

Tarboton, D. G., Ames, D. P., Horsburgh, J. S., Goodall, J. L., Couch, A., Hooper, R., Bales, J., Wang, S., Castronova, A., Seul, M., Idaszak, R., Li, Z., Dash, P., Black, S., Ramirez, M., Yi, H., Calloway, C., & Cogswell, C. (2024). HydroShare retrospective: Science and technology advances of a comprehensive data and model publication environment for the water science domain. *Environmental Modelling & Software*, *172*, 105902. https://doi.org/10.1016/j.envsoft.2023.105902

Tobler, W. R. (1970). A Computer Movie Simulating Urban Growth in the Detroit Region. *Economic Geography*, *46*, 234. https://doi.org/10.2307/143141

Vandewalle, R. C., Barley, W. C., Padmanabhan, A., Katz, D. S., & Wang, S. (2021). Understanding the multifaceted geospatial software ecosystem: A survey approach. *International Journal of Geographical Information Science*, *35*(11), 2168–2186. https://doi.org/10.1080/13658816.2020.1831514

Vandewalle, R., Kang, J.-Y., Yin, D., & Wang, S. (2019). Integrating CyberGIS-Jupyter and spatial agent-based modelling to evaluate emergency evacuation time. *Proceedings of the 2nd ACM SIGSPATIAL International Workshop on GeoSpatial Simulation*, 28–31. https://doi.org/10.1145/3356470.3365530

Vaswani, A., Shazeer, N., Parmar, N., Uszkoreit, J., Jones, L., Gomez, A. N., Kaiser, L., & Polosukhin, I. (2023). *Attention Is All You Need* (No. arXiv:1706.03762). arXiv. https://doi.org/10.48550/arXiv.1706.03762

Wang, H., Fu, T., Du, Y., Gao, W., Huang, K., Liu, Z., Chandak, P., Liu, S., Van Katwyk, P., & Deac, A. (2023). Scientific discovery in the age of artificial intelligence. *Nature*, *620*(7972), 47–60.

Wang, L., Ma, C., Feng, X., Zhang, Z., Yang, H., Zhang, J., Chen, Z., Tang, J., Chen, X., Lin, Y., Zhao, W. X., Wei, Z., & Wen, J. (2024). A survey on large language model based autonomous agents. *Frontiers of Computer Science*, *18*(6), 186345. https://doi.org/10.1007/s11704-024-40231-1

Wang, Q., Qian, C., Li, X., Yao, Z., Zhou, G., & Shao, H. (2024). *Lens: A Foundation Model for Network Traffic* (No. arXiv:2402.03646). arXiv. https://doi.org/10.48550/arXiv.2402.03646

Wang, S. (2010). A CyberGIS Framework for the Synthesis of Cyberinfrastructure, GIS, and Spatial Analysis. *Annals of the Association of American Geographers*, *100*(3), 535–557. https://doi.org/10.1080/00045601003791243

Wang, S. (2013). CyberGIS: Blueprint for integrated and scalable geospatial software ecosystems. *International Journal of Geographical Information Science*, *27*(11), 2119–2121. https://doi.org/10.1080/13658816.2013.841318

Wang, S., Hu, T., Xiao, H., Li, Y., Zhang, C., Ning, H., Zhu, R., Li, Z., & Ye, X. (2024). GPT, large language models (LLMs) and generative artificial intelligence (GAI) models in geospatial science: A systematic review. *International Journal of Digital Earth*. https://www.tandfonline.com/doi/abs/10.1080/17538947.2024.2353122

Wang, Z., Jiang, R., Xue, H., Salim, F. D., Song, X., Shibasaki, R., Hu, W., & Wang, S. (2024). Learning spatio-temporal dynamics on mobility networks for adaptation to open-world events. *Artificial Intelligence*, *335*, 104120. https://doi.org/10.1016/j.artint.2024.104120

Wei, C., Zhang, Y., Zhao, X., Zeng, Z., Wang, Z., Lin, J., Guan, Q., & Yu, W. (2024). GeoTool-GPT: A trainable method for facilitating Large Language Models to master GIS tools. *International Journal of Geographical Information Science*, *0*(0), 1–25. https://doi.org/10.1080/13658816.2024.2438937

Wilson, D., Lin, X., Longstreet, P., & Sarker, S. (2011). Web 2.0: A Definition, Literature Review, and Directions for Future Research. *AMCIS 2011 Proceedings - All Submissions*. https://aisel.aisnet.org/amcis2011_submissions/368

Wu, T., Luo, L., Li, Y.-F., Pan, S., Vu, T.-T., & Haffari, G. (2024). *Continual Learning for Large Language Models: A Survey* (No. arXiv:2402.01364). arXiv. https://doi.org/10.48550/arXiv.2402.01364

Xia, F. F. (2022). GIS Software Product Development Challenges in the Era of Cloud Computing. In B. Li, X. Shi, A.-X. Zhu, C. Wang, & H. Lin (Eds.), *New Thinking in GIScience* (pp. 129–142). Springer Nature. https://doi.org/10.1007/978-981-19-3816-0_15

Xie, Y., Jia, X., Shekhar, Bao, H., & Zhou, X. (2021). Significant DBSCAN+: Statistically Robust Density-based Clustering. *ACM Transactions on Intelligent Systems and Technology (TIST)*. https://doi.org/10.1145/3474842

Xie, Y., Wang, Z., Mai, G., Li, Y., Jia, X., Gao, S., & Wang, S. (2023). Geo-Foundation Models: Reality, Gaps and Opportunities. *Proceedings of the 31st ACM International Conference on Advances in Geographic Information Systems*, 1–4. https://doi.org/10.1145/3589132.3625616

Xu, H., Han, L., Yang, Q., Li, M., & Srivastava, M. (2024). Penetrative AI: Making LLMs Comprehend the Physical World. *Proceedings of the 25th International Workshop on Mobile Computing Systems and Applications*, 1–7. https://doi.org/10.1145/3638550.3641130

Xu, J., & Tao, R. (2024). Map Reading and Analysis with GPT-4V(ision). *ISPRS International Journal of Geo-Information*, *13*(4), Article 4. https://doi.org/10.3390/ijgi13040127

Yang, C., Goodchild, M., Huang, Q., Nebert, D., Raskin, R., Xu, Y., Bambacus, M., & Fay, D. (2011). Spatial cloud computing: How can the geospatial sciences use and help shape cloud computing? *International Journal of Digital Earth*, *4*(4), 305–329. https://doi.org/10.1080/17538947.2011.587547

Yang, C., Raskin, R., Goodchild, M., & Gahegan, M. (2010). Geospatial Cyberinfrastructure: Past, present and future. *Computers, Environment and Urban Systems*, *34*(4), 264–277. https://doi.org/10.1016/j.compenvurbsys.2010.04.001

Yang, E.-W., & Velazquez-Villarreal, E. (2024). *AI-HOPE: An AI-Driven conversational agent for enhanced clinical and genomic data integration in precision medicine research* (p. 2024.11.27.24318113). medRxiv. https://doi.org/10.1101/2024.11.27.24318113

Yang, J., Wang, Z., Lin, Y., & Zhao, Z. (2024). *Problematic Tokens: Tokenizer Bias in Large Language Models* (No. arXiv:2406.11214). arXiv. https://doi.org/10.48550/arXiv.2406.11214

Zenil, H. (2013). *A computable universe: Understanding and exploring nature as computation*. World Scientific. https://books.google.com/books?hl=en&lr=&id=SGG6CgAAQBAJ&oi=fnd&pg=PR7&dq=+A+Computable+Universe:+Understanding+and+Exploring+Nature+as+Computation&ots=3kqTnvrU7g&sig=n-7EiMbItqJik7umLyqHnxUVbZ8

Zhang, H., Xu, J.-J., Cui, H.-W., Li, L., Yang, Y., Tang, C.-S., & Boers, N. (2024). When Geoscience Meets Foundation Models: Toward a general geoscience artificial intelligence system. *IEEE Geoscience and Remote Sensing Magazine*, 2–41. IEEE Geoscience and Remote Sensing Magazine. https://doi.org/10.1109/MGRS.2024.3496478

Zhang, W., Han, J., Xu, Z., Ni, H., Liu, H., & Xiong, H. (2024). Urban Foundation Models: A Survey. *Proceedings of the 30th ACM SIGKDD Conference on Knowledge Discovery and Data Mining*, 6633–6643. https://doi.org/10.1145/3637528.3671453

Zhang, Y., He, Z., Li, J., Lin, J., Guan, Q., & Yu, W. (2024). MapGPT: An autonomous framework for mapping by integrating large language model and cartographic tools. *Cartography and Geographic Information Science*, *51*(6), 717–743. https://doi.org/10.1080/15230406.2024.2404868

Zhang, Y., Wei, C., He, Z., & Yu, W. (2024). GeoGPT: An assistant for understanding and processing geospatial tasks. *International Journal of Applied Earth Observation and Geoinformation*, *131*, 103976. https://doi.org/10.1016/j.jag.2024.103976

Zhang, Y., Zhang, W., Zeng, Z., Jiang, K., Li, J., Min, W., Luo, W., Guan, Q., Lin, J., & Yu, W. (2025). MapReader: A framework for learning a visual language model for map analysis. *International Journal of Geographical Information Science*, *0*(0), 1–36. https://doi.org/10.1080/13658816.2025.2455112

Zhao, Q., Yu, L., Li, X., Peng, D., Zhang, Y., & Gong, P. (2021). Progress and Trends in the Application of Google Earth and Google Earth Engine. *Remote Sensing*, *13*(18), Article 18. https://doi.org/10.3390/rs13183778

Zheng, J., Shi, C., Cai, X., Li, Q., Zhang, D., Li, C., Yu, D., & Ma, Q. (2025). *Lifelong Learning of Large Language Model based Agents: A Roadmap* (No. arXiv:2501.07278). arXiv. https://doi.org/10.48550/arXiv.2501.07278

Zheng, Z., Ermon, S., Kim, D., Zhang, L., & Zhong, Y. (2024). Changen2: Multi-Temporal Remote Sensing Generative Change Foundation Model. *IEEE Transactions on*

*Pattern Analysis and Machine Intelligence*, 1–17. IEEE Transactions on Pattern Analysis and Machine Intelligence. https://doi.org/10.1109/TPAMI.2024.3475824

Zhu, A.-X., Zhao ,Fang-He, Liang ,Peng, & and Qin, C.-Z. (2021). Next generation of GIS: Must be easy. *Annals of GIS*, *27*(1), 71–86. https://doi.org/10.1080/19475683.2020.1766563